\documentclass{article}

\PassOptionsToPackage{}{natbib}
\usepackage{iclr2025_conference}
\usepackage{times} 
\usepackage[utf8]{inputenc} % allow utf-8 input
\usepackage[T1]{fontenc}    % use 8-bit T1 fonts
\usepackage{hyperref}       % hyperlinks
\usepackage{url}            % simple URL typesetting
\usepackage{booktabs}       % professional-quality tables
\usepackage{amsfonts}       % blackboard math symbols
\usepackage{nicefrac}       % compact symbols for 1/2, etc.
\usepackage{microtype}      % microtypography
\usepackage{wrapfig,lipsum,booktabs}

\usepackage{xcolor}         % colors

\usepackage{bbm}
\usepackage[linesnumbered,ruled,vlined]{algorithm2e}
\usepackage{caption}
\usepackage{float}
\usepackage{listings}
\usepackage{multirow}
\usepackage{subcaption}
\usepackage{tikz}
\usepackage{pgfplots}
\usepackage{verbatimbox}
\usepackage{enumitem}
\usepackage[normalem]{ulem}
\pgfplotsset{compat=1.18}

\usepackage{soul}
\iclrfinalcopy 

\newcommand{\ours}{{\sc LASER}}
\newcommand{\oursbf}{{\sc \textbf{LASER}}}

\newcommand{\sep}[0]{~|~}

\makeatletter
\newcommand{\oeq}{\mathbin{\mathpalette\make@circled{=}}}
\newcommand{\oneq}{\mathbin{\mathpalette\make@circled{\neq}}}
\newcommand{\make@circled}[2]{%
  \ooalign{$\m@th#1\smallbigcirc{#1}$\cr\hidewidth$\m@th#1#2$\hidewidth\cr}%
}
\newcommand{\smallbigcirc}[1]{%
  \vcenter{\hbox{\scalebox{0.73}{$\m@th#1\bigcirc$}}}%
}
\makeatother

\definecolor{sclgreen}{rgb}{0,0.46,0}
\definecolor{sclblue}{rgb}{0.02,0.42,0.74}
\definecolor{scllightgrey}{rgb}{0.94,0.94,0.94}%
\definecolor{sclgreyblue}{rgb}{0.3,0.4,0.6}%
\definecolor{sclcyan}{rgb}{0.1,0.4,0.6}%
\definecolor{sclpurple}{rgb}{0.71,0,0.85}%102, 219, 255)
\definecolor{sclyellow}{rgb}{0.9,0.6,0.05}%255, 144, 33
\definecolor{sclorange}{rgb}{1,0.36,0.03}%255, 144, 33
\definecolor{sclred}{rgb}{0.6,0.2,0.0}%

\SetCommentSty{mycommfont}

\lstdefinelanguage{scallop}{
    keywords={import,type,const,rel,query,usize,where,as,String,i8,i32,i64,usize,u8,u16,u32,u64},keywordstyle=\color{blue},%
    morekeywords=[2]{and,or,not,implies,==,+,-,*,/},keywordstyle=[2]\color{sclpurple},%
    morekeywords=[3]{count,sum,prod,min,max,exists,forall,unique,top,categorical,uniform},keywordstyle=[3]\color{sclorange},%
    morecomment=[s]{/*}{*/},%
    commentstyle=\color{sclgreen},%
    morecomment=[l]{//},%
    morestring=[b]",stringstyle=\color{sclyellow}
}

\lstdefinelanguage{mypython}{
    keywords={class,def,str,return,if,elif,else,for,in,while,int,List,Tuple},keywordstyle=\color{blue},%
    morekeywords=[2]{self},
    keywordstyle=[2]\color{sclred},
    morekeywords=[3]{__init__},
    keywordstyle=[3]\color{sclcyan},
    morecomment=[s]{"""}{"""},commentstyle=\color{sclgreen},%
    morecomment=[l]{\#},%
    morestring=[b]",stringstyle=\color{sclorange}
}

\SetKwInput{KwInput}{Input}
\SetKwInput{KwOutput}{Output}

\definecolor{mygreen}{HTML}{D5E8D4}
\definecolor{myred}{HTML}{F8CECC}
\definecolor{myorange}{HTML}{ffcf99}
\definecolor{myblue}{HTML}{99c8f2}
\definecolor{mypurple}{HTML}{E1D5E7}
\definecolor{myyellow}{HTML}{FFF2CC}

\definecolor{mydeepgreen}{HTML}{82B366}
\definecolor{mydeepred}{HTML}{B85450}
\definecolor{mydeeporange}{HTML}{e38820}
\definecolor{mydeepblue}{HTML}{408bcf}
\definecolor{mydeeppurple}{HTML}{9673A6}
\definecolor{mydeepyellow}{HTML}{D6B656}

\definecolor{mylightblue}{HTML}{cfe2fa}
\definecolor{mylightorange}{HTML}{faeccf}

\usepackage{latexsym}
\usepackage{amsmath}
\usepackage{amssymb}
\usepackage{bm}
\usepackage{mathrsfs}
\usepackage{url}

\newcommand{\Dcal}{\mathcal{D}}

\newcommand{\Lcal}{\mathcal{L}}

\newcommand{\Pcal}{\mathcal{P}}

\ifx\BlackBox\undefined
\newcommand{\BlackBox}{\rule{1.5ex}{1.5ex}}  % end of proof
\fi

\ifx\QED\undefined
\def\QED{~\rule[-1pt]{5pt}{5pt}\par\medskip}
\fi

\ifx\proof\undefined
\newenvironment{proof}{\par\noindent{\bf Proof\ }}{\hfill\BlackBox\\}
\fi

\ifx\theorem\undefined
\newtheorem{theorem}{Theorem}

\fi

\ifx\axiom\undefined
\newtheorem{axiom}[theorem]{Axiom}
\fi

\newcommand{\secref}[1]{Section~\ref{#1}}

\newcommand{\appref}[1]{Appendix~\ref{#1}}

\newcommand{\figref}[1]{Figure~\ref{#1}}
\newcommand{\tabref}[1]{Table~\ref{#1}}

\renewcommand{\url}[1]{{\sffamily #1}}

\newcommand{\anglebr}[1]{\langle #1 \rangle}

\title{\ours: A Neuro-Symbolic Framework for Learning Spatio-Temporal Scene Graphs with Weak Supervision}

\author{%
  ~~~~~\textbf{Jiani Huang\textsuperscript{§}~~~~~~~~~~~~~~~~~~~~Ziyang Li\textsuperscript{§}~~~~~~~~~~~~~~~~~~~~Mayur Naik\textsuperscript{§}~~~~~~~~~~~~~~~~~~~~Ser-Nam Lim\textsuperscript{†}} \\
  ~~~~~~~~~~\texttt{\{jianih, liby99, mhnaik\}@seas.upenn.edu}~~~~~~~~~~~~~~~~~~~~\texttt{sernam@ucf.edu} \\
~~~~~~~~~~~~~~~~~~~~~~~~~\textsuperscript{§}University of Pennsylvania~~~~~~~~~~~~~~~\textsuperscript{†}University of Central Florida
}

\begin{document}

\maketitle
\begin{abstract}
%Human annotation of spatio-temporal scene graphs (STSG) in videos is highly labor-intensive, which constrains the scale of datasets available for fully supervised training of STSG generators. 
%To address this, we aim to utilize readily available video captions as weak supervision signals, for which we introduce \ours, a neuro-symbolic framework for training STSG generators using video captions. 
Supervised approaches for learning spatio-temporal scene graphs (STSG) from video are greatly hindered due to their reliance on STSG-annotated videos, which are labor-intensive to construct at scale.
Is it feasible to instead use readily available video captions as weak supervision?
To address this question, we propose \ours, a neuro-symbolic framework to enable training STSG generators using only video captions. 
\ours\ employs large language models to first extract logical specifications with rich spatio-temporal semantic information from video captions.
\ours\ then trains the underlying STSG generator to align the predicted STSG with the specification.
The alignment algorithm overcomes the challenges of weak supervision by leveraging a differentiable symbolic reasoner and using a combination of contrastive, temporal, and semantics losses.
The overall approach efficiently trains low-level perception models to extract a fine-grained STSG that conforms to the video caption.
In doing so, it enables a novel methodology for learning STSGs without tedious annotations.
% We evaluate LASER on three video datasets—OpenPVSG, 20BN-Something-Something, and MUGEN—achieving significant improvements over baselines, including a 12.65\% accuracy increase on OpenPVSG, 7\% on 20BN-Something-Something, and 5.2\% on MUGEN.
% \mayur{Fix rest of abstract.}
We evaluate our method on three video datasets: OpenPVSG, 20BN, and MUGEN.
% We demonstrate that our method achieves superior accuracy compared to existing baselines, exceeding performance by over 12.65\% on the OpenPVSG dataset \todo{change the tone}, outperforming the baseline in 71\% of predicate predictions on the 20BN-Something-Something dataset, and showing a 5.2\% improvement on the MUGEN dataset.
% On OpenPVSG, a real life video caption dataset, our approach demonstrates significant improvements, achieving a Recall@1 Unary accuracy of 0.2778 on OpenPVSG compared to the best performing baseline, IPS Conv's 0.1513, with notable gains in higher recall levels. 
Our approach demonstrates substantial improvements over \textit{fully-supervised} baselines. On OpenPVSG, \ours~achieves a unary predicate prediction accuracy of $27.78\%~(+12.65\%)$ and a binary recall@5 of $0.42~(+0.22)$.  
Furthermore, \ours\ exceeds baselines by $7\%$ on 20BN and $5.2\%$ on MUGEN in terms of overall predicate prediction accuracy.

\end{abstract}

\vspace{-10px}
\section{Introduction}

% \mayurcomment{Global: Citations appear as "... author (year)". Separate from surrounding text: "... (author, year)".}

\begin{figure*}[t]
    \centering
    \includegraphics[width=\textwidth]{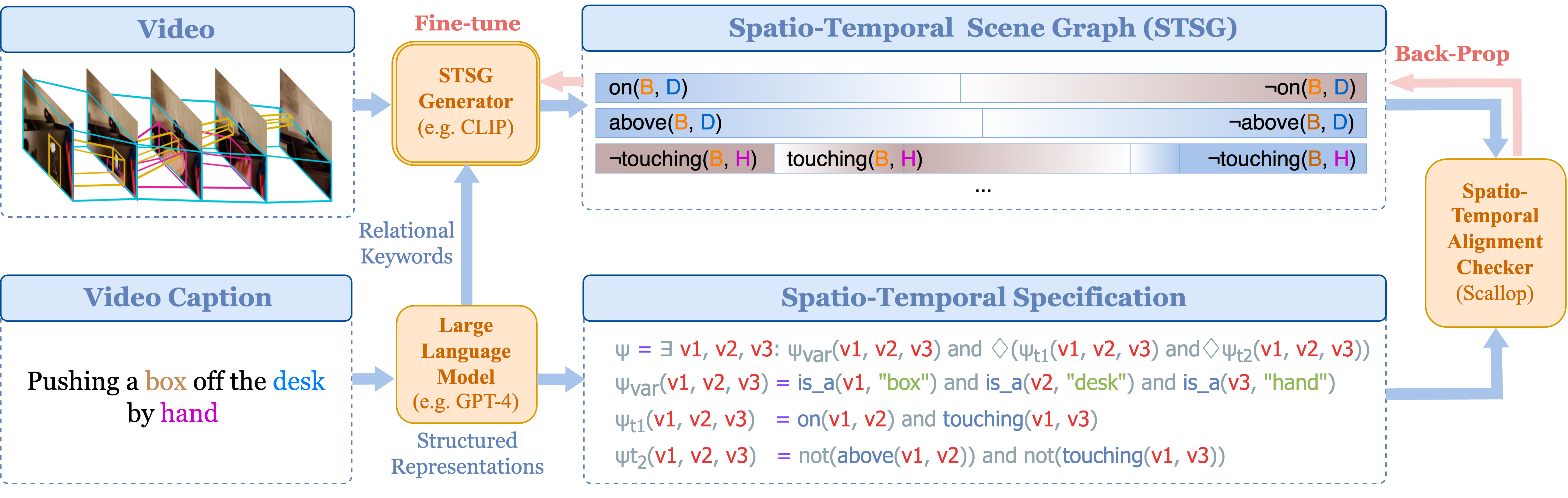}
    \caption{
%     \mayurcomment{Replace Figure 1 with this. Also try to make spec less mathematical:
% 1) $\wedge$ -> {\it and}, and $\neg$ -> {\it not};
% 2) Write it top-down: remove $\psi$, make the last line the first line;
% 3) and then say ``where $\psi_{var}$ = ... and $\psi_{t1}$ = ... and $\psi_{t2}$ = ...";
% 4) say in caption what the diamond operator is.}  
Illustration of the learning pipeline of \ours.
The goal is to fine-tune a vision-language model to produce STSG without direct supervision on ground truth STSG labels.
\ours~relies on video captions for weak-supervision labels.
We apply an LLM to extract a spatio-temporal specification from video captions.
The LLM-inferred relational keywords, along with the input video, are then passed to a vision-language model to generate an STSG.
At the end, a spatio-temporal alignment checker uses the specification to derive an alignment loss, capturing issues in the predicted STSG.
The differentiable checker effectively back-propagates the loss to the vision-language model.
}
    \label{fig:learning_pipeline}
    \vspace{-0.15in}
\end{figure*}

Understanding video semantics has gained prominence due to a wide range of applications such as video search, text-video retrieval, video question answering, video segmentation, and video captioning.
Video semantics constitutes two crucial aspects:
\textit{spatial semantics}, which concern the entities in the video, their individual attributes, and their semantic relationships; and \textit{temporal semantics}, which capture actions and properties evolving through time.
% \mayurcomment{Replace following sentences with new example in Figure 1.}
For example, the video described in Figure \ref{fig:learning_pipeline} by the phrase \textit{``pushing a box off the desk by hand''} involves entities like \textit{``box''} and \textit{``hand''}, which are connected by the spatial relation \textit{``touching''}.
It also features two temporally consecutive states: the \textit{``box''} is first \textit{``on''} the \textit{``desk''}, and then \textit{``not above''} the \textit{``desk''}.

To explicitly learn combined spatial and temporal semantics, a structured representation called \textit{Spatio-Temporal Scene Graph} (STSG) \citep{shang2017video, zhu2022scene} has been proposed
to represent entity relations throughout a video. 
Existing approaches for learning STSG from video data are typically fully-supervised, e.g.,   
\cite{nag2023unbiased, Cong2021sttran}.
They can potentially learn high-fidelity STSGs from video data but are greatly hindered in practice due to the complexity of low-level annotations that are laborious to obtain \citep{yang2023panoptic}.

% It is particularly challenging to use video captions since we need a common-ground to meaningfully compare the spatial and temporal information between the captions and the predicted STSGs. 
% To resolve this, we propose transforming captions into \textit{logical specifications}, which can then be systematically checked against the STSG for semantic alignment.

Weak supervision emerges as a promising approach to address this challenge. 
For example, the vast availability of video captions provides a valuable source of weak supervisory signals. 
However, key difficulties arise in effectively learning STSGs from such weak supervision. 
Is it even feasible to use video captions given the sparsity and noise in the signals they provide? 
Captions often focus only on the primary objects, ignoring underlying details, and many temporal signals are either hidden or must be inferred. 
How can we provide useful fine-grained signals under such circumstances? 
To address these challenges, we propose transforming captions into \textit{logical specifications} using large language models to explicitly reveal the hidden spatial and temporal information. 
This transformation creates a shared foundation to systematically align captions with predicted STSGs. 
The alignment process should
a) capture both spatial and temporal nuances to provide fine-grained supervision for underlying STSG generators;
b) allow diversity, naturalness, and fuzziness in the video and caption data; and
c) account for common-sense knowledge that may be implicit or ambiguous in the captions.

We set out by designing STSL, a general and expressive \underline{S}patio-\underline{T}emporal \underline{S}pecification \underline{L}anguage for specifying fine-grained spatio-temporal properties.
STSL is grounded in \textit{Finite Linear Temporal Logic} (LTL$_f$) \citep{de2013linear} which is used to describe temporal properties over finite traces of action and states.
STSL subsumes action sequences commonly seen in video-action alignment tasks \citep{chang2019d3tw} while capturing additional temporal nuances such as ``until'' ($\mathbf{U}$) and ``finally'' ($\lozenge$). 
It also allows to express common-sense constraints
for extra supervision.
Finally, combined with relational predicates extracted from natural language, such as ``is pushing off'' and ``lies above'', it can even specify the open-domain spatial semantics of videos.

We now introduce \ours~(\underline{L}earning to \underline{A}lign for \underline{S}patio-t\underline{E}mporal \underline{R}epresentations), a novel framework to enable training STSG generators using only video captions.
As illustrated in Figure~\ref{fig:learning_pipeline}, \ours~enhances a vision-language model by aligning its predicted STSGs with STSL specifications derived from video captions using large language models. 
This alignment process is carried out in a divide-and-conquer fashion, where the caption is broken down into temporally related events, each of which must correspond with a portion of the STSG. 
We enhance the alignment process in two important aspects. 
First, to ensure precise optimization, we implement a neuro-symbolic alignment checker atop the Scallop framework \citep{li2023scallop}, making the alignment both probabilistic and differentiable.
This enables seamless integration into an end-to-end learning pipeline. 
Second, to complement the weak supervision, we introduce a multi-faceted loss function that includes contrastive, temporal, and semantic components, which provide additional layers of supervision.

We conduct an extensive evaluation of LASER across various dimensions, demonstrating its broad utility and effectiveness. 
Being model-agnostic, \ours~is capable of fine-tuning a wide range of STSG generators, including open-world generators in our evaluation such as CLIP \citep{radford2021clip}, VIOLET \citep{fu2021violet}, and SigLIP \citep{zhai2023sigmoid}, as well as closed-world generators like MLP classifiers for STSG generation. 
To evaluate its versatility, we apply LASER to diverse datasets such as OpenPVSG \citep{yang2023panoptic}, 20BN \citep{goyal201720bn}, and MUGEN \citep{hayes2022mugen}, which encompass open-domain vocabularies, specifications with varied patterns, and both synthetic and complex real-world videos.
%OpenPVSG, our most challenging dataset, is comprised of complex real-world videos and captions. 
Despite challenges such as an excessive number of relevant entities, diverse open-domain labels, and fuzzy captions, \ours~consistently outperforms even fully supervised baselines in STSG generation on all three datasets.
To further evaluate the data efficiency of \ours, we train the STSG generator with just $10\%$ of the training data, resulting in an average of $70.75\%$ of the performance gains achieved with the full dataset.

We summarize the main contributions of this work as follows: 
\begin{enumerate}[topsep=0pt,itemsep=-1ex,partopsep=1ex,parsep=1ex,leftmargin=0.5cm]
    \item We introduce a novel formulation of spatio-temporal scene graph learning as a weakly supervised task driven by video captions.
    \item We design STSL, a general and expressive spatio-temporal specification language, for specifying fine-grained video semantics.
    \item We implement a differentiable neuro-symbolic alignment checker to relate between an STSL specification and a spatio-temporal scene graph.
    % for specifying desired STSG behavior, along with its checker implementation in Scallop.
    % \item We provide a generic natural language to STSL program prompt template utilizing GPT-4.
    \item We propose \ours, a model-agnostic, end-to-end differentiable framework for learning spatio-temporal scene graphs with weak supervision from video captions.
    % \item We design a loss function combining alignment loss, temporal constraint semantic loss, and contrastive loss for specification-guided learning.
    \item We empirically evaluate \ours~on three video understanding datasets, demonstrating superior performance in video semantics extraction tasks. 
    % Compared to fully supervised methods, we achieve  $0.42$ VIOU improvement on OpenPVSG, $0.03$ F1-score improvement on 20BN, and $5.2\%$ accuracy improvement on MUGEN datasets, using only a fraction of the training labels.
\end{enumerate}

\vspace{-5px}
\section{Related Work}
\vspace{-5px}

\textbf{Structured Representation of Image/Video Semantics.}
Significant advances have been made in representing structured information within vision data. 
A widely adopted representation for capturing spatial semantics in images is \textit{Scene Graph} \citep{kuznetsova2018imagev4, lu2016visual}, with various generation techniques emerging over the years \citep{zhu2022scene, Liu2021convsgg, huang2020referexpr, huang2021scallop, yang2018graph, vieira2024li}. 
More recently, research has increasingly focused on extending these representations to integrate both spatial and temporal semantics in videos \citep{yang2023panoptic, li2022essgg}.
However, learning spatio-temporal structures remains a challenging open problem, which \ours{} addresses by introducing a neuro-symbolic approach.

% \mayurcomment{Add a para to make this section start less abruptly. Start by talking about scene graphs from 1nd para of intro ("Significant progress has been made ...").  Then give a 1 sentence overview of the related work that makes the following two paras natural (e.g. Existing approaches can be broadly classified into fully supervised and unsupervised).}

\textbf{Video Scene Graph Learning.} 
Learning video scene graphs has attracted significant attention from the vision community. 
Various tasks, including entity tracking, object identification, dynamic relation analysis, and pathfinding, are being explored \citep{sun2023trace, xu2022meta, shang2017video}. 
New techniques involving spatio-temporal aware networks have been developed \citep{nag2023unbiased, Cong2021sttran, ji2021detecting, sun2019videobert}.
% However, learning fine-grained video scene graphs through weak supervision is still under exploration.
A few recent works have developed point-solutions for extracting fine-grained video semantics  \citep{lee2023neuro, apriceno2022neuro, chang2019d3tw}.
Such approaches includes both dynamic time warping  \citep{dvornik2021drop, chang2019d3tw, richard2018neuralnetwork, ding2018weakly}, soft nearest neighbor \citep{han2022temporal, dwibedi2019temp}, and semantic loss \citep{xu2022dont}.
To our knowledge, \ours{} \emph{is the first framework to train STSG generation models using video captions as weak-supervisory labels}.

\textbf{Vision Language Pretraining.}
Vision language pretraining is crucial in video understanding and has a wide range of downstream applications.
Current works have succeeded in learning visual representations using large-scale paired visual-textual data through contrastive learning in both image-text  \citep{zhai2023sigmoid, radford2021clip, jia2021scaling}  and video-text \citep{li2021align, xu2021videoclip, miech2019end} representation learning.
Recent works also explore the viability of utilizing pretrained foundation models for generating image and video scene graphs \citep{shindo2024deisam, liang2024ckt, zhang2023learning, Yao2021ICCV}.

\vspace{-5px}
\section{Methodology}
\vspace{-5px}

We begin by presenting the high-level problem definition.
We are given a dataset $\Dcal$ of video-caption pairs $(X, c)$, where $X = [x_1, \dots, x_n]$ is a video containing $n$ frames, and $c$ is its video caption.
% To provide a more fine grained weak supervision, we consult an external foundation model, GPT-4, to generate and approximate the programmatic spatio-temporal specification $\psi$ for the caption $c$.
% We are given a dataset $\Dcal$ of video-specification pairs $(X, \psi)$, where $X = [x_1, \dots, x_n]$ is a video containing $n$ frames, and $\psi$ is a spatio-temporal specification using LTL$_f$.
% We then convert the caption $c$ into specifications $\psi$,  a spatio-temporal specification in LTL$_f$, utilizing large language model.
We then transform the caption $c$ into a spatio-temporal specification $\psi$ in LTL$_f$ using a large language model.
We wish to learn a neural model $M_\theta$ which extracts a spatio-temporal scene graph $\hat{\mathbf{r}} = M_\theta(X)$ that conforms to the corresponding  specification $\psi$.
During training time, given a loss function $\Lcal$, we aim to minimize the following main objective:
\vspace{-12px}
\begin{figure}[H]
\footnotesize
\begin{align}
J(\theta) &= \textstyle\frac{1}{|\Dcal|} \sum_{(X, \psi)  \in \Dcal} \Lcal(\Pr(\hat{\mathbf{r}} \models \psi ~|~ \theta), 1),
\end{align}
\vspace{-20px}
\end{figure}

\noindent where $\Pr(\hat{\mathbf{r}} \models \psi ~|~ \theta)$ is the alignment score (probability of alignment) computed by our spatio-temporal alignment checker, conditioned on the model parameter $\theta$. 
We illustrate the full learning pipeline in \figref{fig:learning_pipeline} and detail the process in this section. 
We first describe our video semantics representation using a probabilistic relational database (\S \ref{sec:method-sym-db}).
Then, we present our specification language STSL (\S \ref{sec:method-stsl}) and its alignment checker (\S \ref{sec:method-ltl-inference}).
We further demonstrate how to automatically convert natural language captions to a specification in STSL using LLM (\S \ref{sec:method-nl2spec}).
In \S \ref{sec:method-loss}, we present our multi-faceted loss function design comprising contrastive, temporal, and semantic components. 

\subsection{Video to Probabilistic Relational Database}
\label{sec:method-sym-db}

\begin{figure}
    \centering
    \includegraphics[width=.9\linewidth]{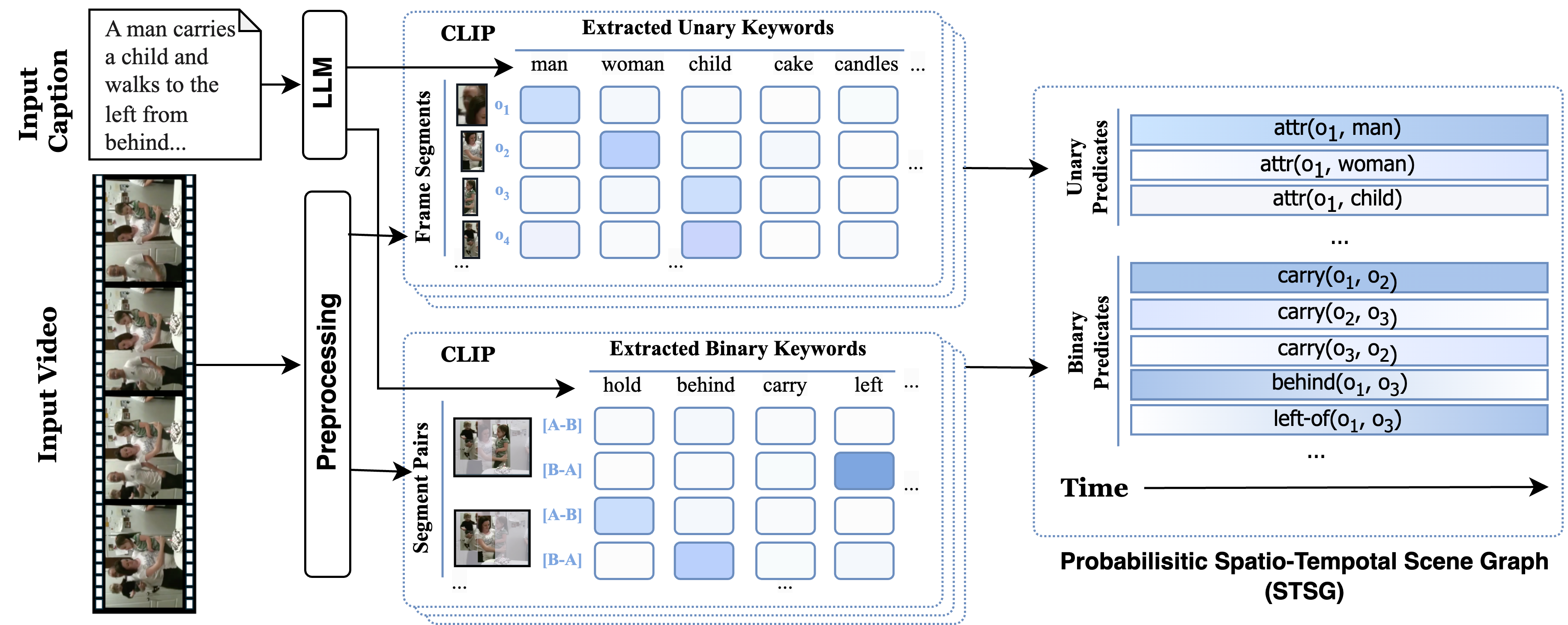}
    \caption{
    % An illustration using CLIP as the backbone model in the pipeline generating Probabilistic STSG generation. 
   Pipeline illustration with CLIP as the backbone model for probabilistic STSG generation.
    % This pipeline generates a differentiable probabilistic STSG from an input video and caption pair. 
    % An LLM first extracts relational keywords from the caption, which are uniformly applied to both the STSG and STSL. 
    % The video is then preprocessed by cropping objects in each frame, aligning them with unary keywords, and using weighted masking on object pairs to distinguish the subject, object, and background, aligning them with binary keywords. 
    % Consistent relationships, such as object categories, are aggregated across the video, yielding a probabilistic STSG—a database of probabilistic facts.    
    }
    \label{fig:model-architecture}
    \vspace{-10px}
\end{figure}

A probabilistic spatio-temporal scene graph is a probabilistic relational database that contains two types of facts denoted by relations \lstinline{unary_atom} and \lstinline{binary_atom}, for unary and binary predicates respectively, each associated with a probability denoting the likelihood that the fact is true.
%unary, of type ({\small {\tt predicate}} : {\small {\tt String}}, {\small {\tt fid}} : {\small {\tt Time}}, {\small {\tt o}} : {\small {\tt ObjID}}), and binary, of type 
%({\small {\tt predicate}} : {\small {\tt String}}, {\small {\tt fid}} : {\small {\tt Time}}, {\small {\tt o1}} : {\small {\tt ObjID}},  {\small {\tt o2}} : {\small {\tt ObjID}}).
%\begin{lstlisting}%[language=scallop,numbers=none,xleftmargin=.03\textwidth]
%type unary_atom(predicate: String, fid: Time, o1: ObjID)
%type binary_atom(predicate: String, fid: Time, o1: ObjID, o2: ObjID)
%\end{lstlisting}
For example, \lstinline{0.05::unary_atom("deformed", 3, e)} means that \textit{``entity \lstinline{e} is unlikely to be deformed at time stamp 3,''} while \lstinline{0.92::binary_atom("push", 10, h, b)} indicates that \textit{``object \lstinline{h} is highly likely to be pushing object \lstinline{b} at time stamp 10.''}
This flexible representation supports the seamless incorporation of unary and binary keywords into the database. The unified probabilistic database enables \ours\ to be model-agnostic, supporting both closed-domain STSG classification models and open-world vision-language models for converting input video data into relational representations.
With a unified formalization, an STSG generator, $M_\theta$, takes in pixel-based raw video data $X$, and generate a distribution of STSGs, which is encoded as a probabilistic relational database $\hat{\mathbf{r}}$.

We illustrate the STSG generation pipeline using an open-world vision-language model, CLIP, in \figref{fig:model-architecture}. The pipeline generates a probabilistic STSG from an input video and caption pair. An LLM first extracts relational keywords from the caption and passes the keywords to the STSG model. The video is then preprocessed by cropping objects in each frame and aligning them with unary keywords, while weighted masking is applied to object pairs to distinguish between subject, object, and background, aligning them with binary keywords. Consistent relationships, such as object categories, are aggregated across the video, resulting in the desired probabilistic STSG.
For closed-domain STSG classification models, the pipeline remains the same in terms of video preprocessing. However, instead of relying on an LLM to extract keywords, these models predict directly over a predefined set of vocabularies.
For brevity, throughout the paper, we use actual unary and binary keywords as predicate names, e.g. \lstinline{deformed(2, e)} and \lstinline{push(10, h, b)}.

% one optimized for generalizability, fine-tuning CLIP \citep{radford2021clip} as the backbone for open-world scene graph generation (\figref{fig:model-architecture}), and the other optimized for accuracy, training a scene graph classifier from scratch (\appref{app:experimental_details}).
% At a high level, the model has two classification heads extracting two types of probabilistic information.
% Note that we assume that a relational schema is pre-defined and given for each task, meaning that the dimensions of classification heads are known beforehand.
% By associating the classification outputs with explicit relational symbols, we then construct the predicted database $\hat{\textbf{r}}$.

% Except for the static properties, others are relational facts that evolve over time, and thus contain the time step number.
% Note that our time step does not directly correspond to each frame.
% While there are other discretizations of time, in this work, each time step represents an ID of a clip, which is a size $k$ sliding window of the video.
% We assume that the STSG is discretized into $m$ time steps.

\subsection{Spatio-Temporal Specification Language (STSL)}
\label{sec:method-stsl}
Linear Temporal Logic (LTL) \citep{Pnueli1977TheTL} is a formal logic system extending propositional logic with concepts about time.
It is commonly used for formally describing temporal events, with applications in software verification \citep{chaki2005concurrent, kesten1998algorithmic} and control \citep{ding2014optimal, sadigh2014learning}.
As we operate on prerecorded, finite-length videos, our language is developed using LTL$_f$ \citep{de2013linear}, which supports LTL reasoning over finite traces.
Thus, we use LTL$_f$ as a framework for specifying events and their temporal relationships.
% Note that LTL$_f$ is a fragment of the more general \textit{Linear Temporal Logic} (LTL) \citep{Pnueli1977TheTL}, which semantics is defined over infinite traces. 
% \mayur{LTL and CTL are incomparable. You probably mean CTL*. See wikipedia.}
% Since this work is centered around pre-recorded and finite-length videos, LTL$_f$ is sufficient for providing temporal specification.

Our STSL (Figures \ref{fig:ltl-high-level-syntax} and \ref{fig:ltl-semantics}) further extends LTL$_f$ by introducing relational predicates and variables.
It starts from the specification $\psi$ which existentially quantifies variables in an STSL formula.
The formula $\varphi$ is inductively defined, with basic elements as relational atoms $\alpha$ of the form $a(t_1, \dots, t_n)$.
Note that the terms $\bar{t} = \{t_1, \dots, t_n\}$ can contain quantified variables to be later grounded into concrete entities based on context $\Gamma$, noted by $\text{subst}_\Gamma(\bar{t})$.
From here, $\varphi$ can be constructed using basic propositional logic components $\land$ (and), $\lor$ (or), and $\neg$ (not).
The system additionally includes temporal unary operators $\square$ (always), $\lozenge$ (finally), $\bigcirc$ (next), and a binary operator $\mathbf{U}$ (until) \citep{albers2009automata}.
% \todo{Update this example with variable }
For example, the description ``A hand continues to touch the box until it drops.'' can be represented as an STSL formula
%\footnotesize
\begin{align}
\label{eqn:touch-and-drop}
\psi = \texttt{touch}(\texttt{h},~\texttt{b}) ~\mathbf{U}~ \texttt{drop}(\texttt{b},~\texttt{\_}).
\end{align}
Note that an argument to the predicate \texttt{drop} is a wildcard (\texttt{\_}), since we do not specify where does the box drops from.
This formula might seem too strict since it requires the two events to be consecutive.
To make the specification more natural, one can change the above formula to ``$\lozenge (\texttt{touch(h, b)} \wedge \lozenge \texttt{drop(b, \_)})$''.
Here, the two events, \texttt{touch} and \texttt{drop}, need to happen in chronological order but are not required to be consecutive.
% Plus, the events need not to start with climbing.

% LTL$_f$ formulas can be evaluated against a concrete finite event sequence for \textit{satisfaction}, or \textit{alignment}.
% Given a space of all possible events $\Sigma$, a size $n$ event sequence $s : \{1 \dots n\} \rightarrow 2^\Sigma$ is a mapping from a time-step $i \in 1 \dots n$ to a set of events happening at that time-step.
% We write $s \models \psi$ iff the sequence $s$ \textit{satisfies} an LTL$_f$ formula $\psi$.
% For example, given the formula $\psi$ from Eqn. \ref{eqn:climb-and-walk}, we have
% $\texttt{[c,c,w,w]} \models \psi$ and
% $\texttt{[c,c,c,c]} \not\models \psi$
% \footnote{We use \texttt{c} and \texttt{w} to denote singleton sets of events containing only \texttt{climb}(\texttt{M}, \texttt{Up}) and \texttt{walk}(\texttt{M}, \texttt{Right}).}.
% Therefore, our choice of neuro-symbolic framework, Scallop, is computationally sufficient for solving this task.
% In this work, we say that a event sequence $s$ is \textit{aligned} with an LTL formula $\psi$ iff $s \models \psi$.
% This computational problem can be expressed in Datalog, the logic programming language supported by Scallop.

\subsection{Natural Language to Programmatic Spatio-Temporal Specification}
\label{sec:method-nl2spec}

\begin{figure}[t!]
    \centering
    \includegraphics[width=\linewidth]{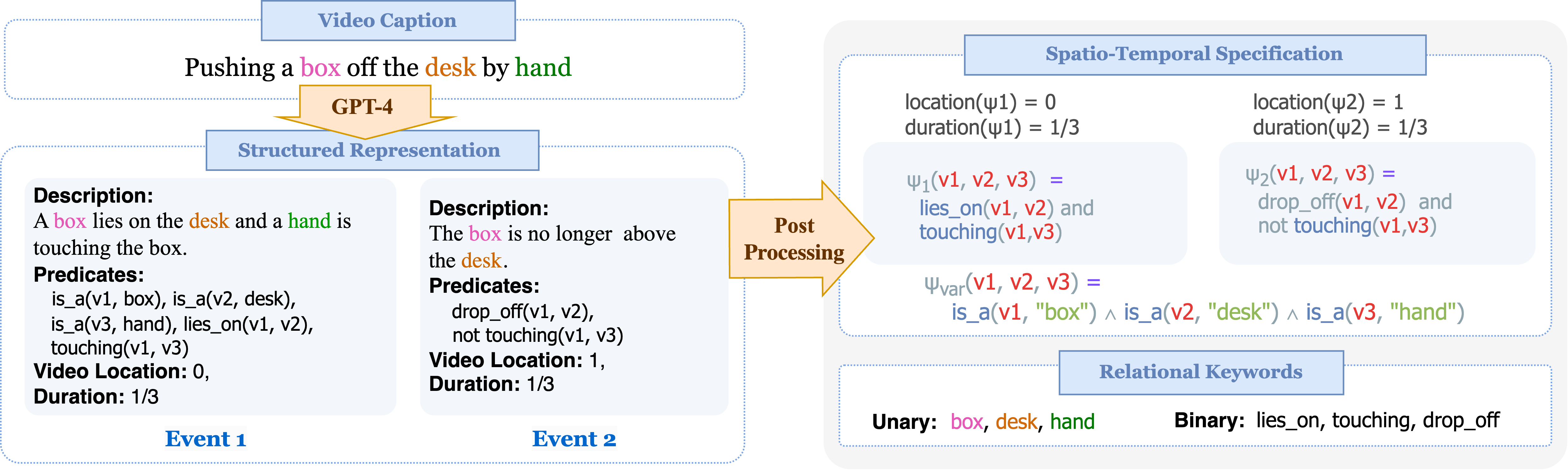}
    \vspace{-10px}
    \caption{
Pipeline utilizing 3-shot GPT-4 to convert natural language captions into: (1) programmatic spatio-temporal specification for alignment score calculation as input to the alignment checker, and (2) unary and binary keywords for predicting the probabilistic STSG as inputs to the neural model.}
    \label{fig:nl2spec}
    \vspace{-10px}
\end{figure}

To leverage the abundance of video captions as weak supervision signals, we employ a large language model (LLM) to automatically extract a programmatic specification $\psi$ from each video caption $c$. 
Directly converting captions into a formal program is particularly challenging for an LLM, especially in a low-data language like STSL. 
We hence use a few-shot learning approach with an LLM to generate an intermediate structured representation of the caption in JSON format.
For each caption $c$, our goal is to convert it into a series of events $\bar{e} = \{e_1, e_2, \dots, e_{n}\}$. 
Each event includes
(a) a detailed natural language description of the event, which guides the generation of subsequent details,
(b) a series of unary, binary, positive, and negative predicates describing the semantics of the scenario,
(c) the location of the event, $\text{loc}(e_i)$, in the video where the event occurs, represented as a fraction of the video length, and
(d) the duration of the event, $\text{dur}(e_i)$, also expressed as a fraction.
In \secref{sec:method-loss}, we explain how these structured representations are incorporated into the loss function.

To extract such structured representations from the caption,
we designed a generic prompt template, which consists of the following components:
(a) examples for temporal specification in fraction numbers: ``0'', ``1/2'', ``2/3'', ``1''.
(b) scene graph keywords, such as object names and relations.
(c) few-shot examples of caption and JSON structured representations pairs.
We illustrate a caption and its structured representation in \figref{fig:nl2spec}, and the full prompt in the appendix.

The programmatic spatio-temporal specification is then generated by postprocessing the events in sequential order. 
Consequently, we can generate the programmatic spatio-temporal specification $\psi$ for the caption as a sequence of events in chronological order:

\vspace{-15px}
\begin{figure}[H]
\footnotesize
\begin{align}
\textstyle
\psi = \lozenge_{e_i \in \bar{e}} \psi_i, \quad \psi_i = \bigwedge_{\phi_j \in \psi_i} \phi_j.
\end{align}
\vspace{-15px}
\end{figure}

\subsection{Spatio-Temporal Alignment Checking}
\label{sec:method-ltl-inference}
\begin{figure}[!t]
\footnotesize
\begin{minipage}{\linewidth}
    \begin{minipage}{0.52\linewidth}
        \begin{minipage}{\linewidth}
            \begin{equation*}
            \arraycolsep=1.4pt
            \begin{array}{rrrl}
                \text{(Formula)} & \varphi & ::= &
                a(\overline{t})
                \sep \varphi_1 \wedge \varphi_2 
                \sep \varphi_1 \vee \varphi_2
                \sep \neg \varphi \\
                & & \sep & 
                \bigcirc \varphi
                \sep \varphi_1 ~\mathbf{U}~ \varphi_2
                \sep \square \varphi
                \sep \lozenge \varphi \\
                \text{(Specification)} & \psi & ::= & \exists v_1, \dots, v_k, \text{s.t.~} \varphi \\
            \end{array}
            \end{equation*}
            \vspace{-5px}
            \captionof{figure}{
            The formal syntax of STSL. 
            Here, $\wedge$, $\vee$, and $\neg$ represents logical ``and'', ``or'', and ``not''.
            Formula may also contain temporal operators $\bigcirc$ (next), $\mathbf{U}$ (until), $\square$ (global), and $\lozenge$ (finally).}
            \label{fig:ltl-high-level-syntax}
        \end{minipage}
        \vspace{5px}
        \begin{minipage}{\linewidth}
            \begin{equation*}
            \arraycolsep=1.4pt
            \begin{array}{rrlcl}
            % \anglebr{w, s} & \models & \psi & \text{iff} & \exists \Gamma, \anglebr{\Gamma, w, s}  \models  \varphi \\
            % \anglebr{\Gamma, w, s} & \models & a(\bar{t}) & \text{iff} & a(\bar{c}) \in w[s] ~\wedge~ \bar{c} = \text{subst}_\Gamma(\bar{t})\\
            % \anglebr{\Gamma, w, s} & \models & \varphi_1 \wedge \varphi_2 & \text{iff} & \anglebr{\Gamma, w, s} \models \varphi_1 ~\wedge \\
            % & & & & \anglebr{\Gamma, w, s} \models \varphi_2 \\
            % \anglebr{\Gamma, w, s} & \models & \neg \varphi & \text{iff} & \anglebr{\Gamma, w, s} \not\models \varphi \\
            % \anglebr{\Gamma, w, s}& \models & \bigcirc \varphi & \text{iff} & \anglebr {\Gamma, w, s + 1} \models \varphi \\
            % \anglebr{\Gamma, w, s}  & \models & \varphi_1 \mathbf{U} \varphi_2 & \text{iff} & \exists i. s \leq i  \land \anglebr {\Gamma, w, i} \models \varphi_2 \\
            % & & & & \forall k. s \leq k < i, \anglebr{\Gamma,  w, k} \models \varphi_1 \\ \\
            \anglebr{w, [s: t]} & \models & \psi & \text{iff} & \psi = \exists v_1, \dots, v_k: \varphi ~\text{and} \\
            & & & & \text{there exists}~ \Gamma : V \mapsto E, \\
            & & & & \langle w, [s:t], \Gamma \rangle \models \varphi \\
            \langle w, [s : t], \Gamma \rangle & \models & \alpha & \text{iff} & \forall i \in s \dots t, \\
            & & & & \text{substitute}_\Gamma(\alpha) \in w[i] \\
            \langle w, [s : t], \Gamma \rangle & \models & \varphi_1 \wedge \varphi_2 & \text{iff} & \langle w, [s : t], \Gamma \rangle \models \varphi_1 ~\text{and}~ \\
            & & & & \langle w, [s : t], \Gamma \rangle \models \varphi_2 \\
            \langle w, [s : t], \Gamma \rangle & \models & \neg \varphi & \text{iff} & \langle w, [s : t], \Gamma \rangle \not\models \varphi \\
            \langle w, [s : t], \Gamma \rangle & \models & \bigcirc \varphi & \text{iff} & \langle w, [s + 1 : t], \Gamma \rangle \models \varphi \\
            \langle w, [s : t], \Gamma \rangle & \models & \square \varphi & \text{iff} & \forall i \in s .. t, \langle w, [i : i], \Gamma \rangle \models \varphi \\
            \langle w, [s : t], \Gamma \rangle & \models & \lozenge \varphi & \text{iff} & \exists i \in s .. t, \langle w, [i : i], \Gamma \rangle \models \varphi \\
            \langle w, [s : t], \Gamma \rangle & \models & \varphi_1 \mathbf{U} \varphi_2 & \text{iff} & \exists i, s < i < t, \\
            & & & & \langle w, [s : i - 1], \Gamma \rangle \models \varphi_1, ~\text{and}~ \\
            & & & & \langle w, [i : i], \Gamma \rangle \models \varphi_2 \\
            \end{array}
            \end{equation*}
            \vspace{-10px}
            \captionof{figure}{
            Formal semantics of STSL.
            $\anglebr{w, [s: t]} \models \psi$ means the STSL specification $\psi$ is \textit{aligned} with the ST-SG $w$ starting from time $s$ till time $t$.
            % The notation $\textbf{w}[i]$ is used to fetch all the facts corresponding to time step $i$ in the world $\textbf{w}$.
            % Note that additional operations of $\vee$ (or), $\square$ (global) and $\lozenge$ (finally) can be represented as a syntax sugar of the above operations.
            We use $w \models \psi$ as an abbreviation for  $\anglebr{w, [1: m]} \models \psi$, where $m$ is the full video length.
            }
            \label{fig:ltl-semantics}
        \end{minipage}
    \end{minipage}
    \hfill
    \begin{minipage}{0.45\linewidth}
        \includegraphics[width=\linewidth]{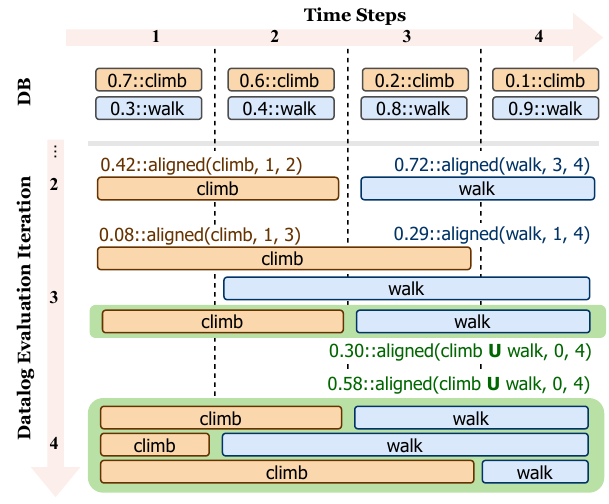}
        \vspace{-10px}
        \captionof{figure}{
        The evaluation process aligning a spatio-temporal scene graph (DB) with a specification  $\texttt{touch}~\mathbf{U}~\texttt{drop}$.
         This figure omits the predicate arguments, concentrating solely on matching sequential events.
         }
        \label{fig:ltl-semantic-align}
    \end{minipage}
\end{minipage}
\vspace{-10px}
\end{figure}

Given a probabilistic database $\mathbf{r}$ that encodes a distribution of STSGs (\S \ref{sec:method-sym-db}), and a specification $\psi$ in STSL, we aim to measure the alignment score $\Pr(\mathbf{r} \models \psi)$ in an end-to-end and differentiable manner.
Conceptually, each probabilistic fact $f$ in the database can be toggled on or off, resulting in $2^{|\mathbf{r}|}$ distinct \textit{worlds}.
Denoting each world (i.e. a discrete STSG) as $w \in \Pcal(\mathbf{r})$, where $\Pcal$ represents power-set, we can check whether the world $w$ satisfies the specification $\psi$ or not (Figure~\ref{fig:ltl-semantics}).
From here, the alignment score can be computed as the sum of the probabilities of worlds satisfying $\psi$:
\begin{equation}
    \footnotesize
    \Pr(\mathbf{r} \models \psi) = \textstyle\sum_{w \in \Pcal(\mathbf{r})} \Pr(w) \cdot \mathbbm{1}[w \models \psi], \quad
    \Pr(w) = \textstyle\prod_{f \in w} \Pr(f) \textstyle\prod_{f' \in \mathbf{r} \backslash w} (1 - \Pr(f'))
\end{equation}
Enumerating all possible worlds is intractable due to its exponential complexity.
Existing general-purpose neuro-symbolic systems like Scallop \citep{li2023scallop} employ scalable algorithms to approximate this probability and greatly reduce the probabilistic reasoning time.
We also note that some of the STSG $w$ sampled from $\Pcal(\mathbf{r})$ might be infeasible due to involving conflicting facts (e.g., a box is above and below a desk at the same time).
To further enhance the logic deduction efficiency, we extend Scallop's ``top-$k$ proofs'' provenance to support general disjunctive constraints and early removal of
infeasible STSGs that do not satisfy the specification.

\ours\ implements the alignment checker using our extended version of Scallop.
% This helps us to succinctly and precisely encode the formal semantics of STSL.
It inductively computes the alignment between a temporal slice of $\textbf{r}$ and an STSL formula.
The whole specification $\psi$ is aligned if the full $\textbf{r}$  satisfies $\psi$ with a concrete variable grounding $\Gamma$, which maps variables to concrete entities.
We also extend Scallop with a variable assignment interface enumerating $\Gamma$ for constraint solving.
% Using Scallop comes with a cost of approximating the probability of alignment.
We illustrate one simplified evaluation process in \figref{fig:ltl-semantic-align}.
The checker iteratively aligns the predicted probabilistic events (simplified to just \texttt{touch} and \texttt{drop}) with the specification.
At the $4$th iteration, $4$ different satisfying alignments are derived, yielding a final aggregated alignment score of $0.9916$.
% Aggregating them together gives the final alignment score $0.58$.

% \input{figures/temporal_match_procedure}
\subsection{Loss Function}
\label{sec:method-loss}

\textbf{Contrastive Learning.}
Unavoidable dataset biases exist in the specification. Contrastive learning can effectively reduce the bias and generate explanations of better quality. 
Let $(X_i, \psi_i)$ and $(X_j, \psi_j)$ be two datapoints in a mini-batch $B$, where $\psi_i$ and $\psi_j$ are the specifications for video $X_i$ and $X_j$ correspondingly. 
If $X_i \models \psi_j$, then it is an extra positive sample to the video $X_i$; otherwise, it is a negative sample to $X_i$.
We can thus define our per-batch contrastive loss $\mathcal{L}_c(B)$:
\begin{align}
\Lcal_c(B) &= \textstyle\frac{1}{|B|^2} 
    \sum_{(X_i, \psi_i) \in B} 
        \sum_{ \substack{(X_j, \psi_j) \in B}} 
            \Lcal(Pr(M_\theta(X_i) \models \psi_j),  \mathbbm{1}[\psi_i =\psi_j])
\label{eqn:time_span}
\end{align}
\vspace{-10px}

% \noindent Note that, due to the one-to-many property existing in the specification-video pairs, it is possible to incorporate noises under the weak supervision setup, as the ground truth symbolic representation of the video is not available to check against the specifications. 

\textbf{Time-Span Supervision.}
A video caption is expanded into a sequence of events using LLM, with each event assigned a specific temporal target, detailing its location and duration within the video, as illustrated in Section \ref{sec:method-nl2spec}. 
By aligning the spatio-temporal specification $\psi$ with the video, we can identify when its sub-specifications, $\psi_1, \psi_2, \dots, \psi_n$, are met. 
This alignment facilitates weak supervision across the entire time span.
We define $\sigma(s, l, d) \in [0, 1]$, the time span alignment score, as a function on actual time stamp $s$, expected time stamp $l$, and expected event duration $d$.
In particular, $\sigma(s, l, d)$ should peak at 1 when the event happens exactly at the expected locations ($s = l$).
In practice, we embed $\sigma$ into the computation of probabilistic alignment between STSG $w$ and an atomic specification $a(\bar{t})$, where we utilize the expected location $\text{loc}(a)$ and the duration $\text{dur}(a)$:
\begin{align}
\sigma(s, l, d) 
&= 
\text{max}(0, 1 - \textstyle\frac{2 |s - l|}{d}),
\\
\Pr(\anglebr{\Gamma, w, s} \models a(\bar{t}))
&=
\Pr(a(\bar{c}) ~|~ a(\bar{c}) \in w[s] ~\wedge~ \bar{c} = \text{subst}_\Gamma(\bar{t})) \cdot \sigma(s, \text{loc}(a), \text{dur}(a)).
\end{align}

\textbf{Semantic Loss.}
To provide further supervision, we resort to human knowledge encoded in the form of integrity constraints. 
We introduce semantic loss reflecting the probability of violating the integrity constraints.
For example, an entity in a video cannot be \texttt{open} and \texttt{closed} at the same time;
an entity that is not \texttt{bendable} cannot be \texttt{deformed}.
These integrity constraints may interweave so heavily that it is hard to use a simple disjoint multi-class classifier to enforce.
We encode all integrity constraints in the form of first-order logic rules, and our reasoning engine generates the probability that these constraints are violated.
We thus have the per-sample semantic loss $\Lcal_s(X_i)$ as an extra-weighted term after calculating the other loss components.
Let $n$ be the number of integrity constraints and let $\texttt{IC}_i$ be the $i$-th integrity constraint, we have

\vspace{-8px}
\begin{figure}[H]
\footnotesize
\begin{align}
\Lcal_s(X_i) = \sum_{i=1}^n \Lcal(\Pr(M_\theta(X_i) \not\models \psi_{\texttt{IC}_i}), 0).
\end{align}
\vspace{-25px}
\end{figure}
\noindent

% Pr(x) = 1 - Pr(!x)
% Pr(w |/= )
% $w |= \psi == ! w |/= \psi$
\vspace{-10px}
\section{Evaluation}
\vspace{-5px}

\label{sec:evaluation}
% Our evaluation aims to answer the following research questions:
% \begin{enumerate}[topsep=0pt,itemsep=-1ex,partopsep=1ex,parsep=1ex,leftmargin=0cm, label={}]
% \item \textbf{RQ1:} How does the accuracy of STSGs learned with weak supervision using LASER compare to baseline methods, including both strongly and weakly supervised models? 
% \item \textbf{RQ2:} Is the STSL language we designed expressive and generic enough for different use cases?
% \item \textbf{RQ3:} How do different components of LASER's loss function contribute to its learning performance?
% \end{enumerate}

% \begin{enumerate}[topsep=0pt,itemsep=-1ex,partopsep=1ex,parsep=1ex,leftmargin=0cm, label={}]
% \item \textbf{RQ1:} Is \ours~effective in train STSG generators?
% \item \textbf{RQ2:} Compared against fully supervised baselines, 
% \item \textbf{RQ3:} Is STSL versatile and expressive? 
% \item \textbf{RQ4:} Is the multifaceted design of \ours~'s loss function necessary?
% \end{enumerate}

% Our evaluation explores several critical aspects of \ours. 
% First, we assess \ours's effectiveness in learning STSG generators. 
% We compare \ours~performance against fully supervised baselines to determine how well it performs in comparison to existing methods. 
% We also investigate the versatility and expressiveness of STSL by applying it across diverse specification patterns. 
% Lastly, we evaluate the multifaceted design of \ours's loss function, assessing the necessity of each component.

Our evaluation attempts to answer several key questions about \ours. 
How effective is \ours\ in learning STSG generators? 
How does its performance compare to fully supervised baselines and existing methods? 
Is STSL versatile and expressive enough when applied to diverse specification patterns? Finally, how essential is the multifaceted design of \ours's loss function?

To address these issues, 
we evaluate \ours~on three datasets: OpenPVSG  \citep{yang2023panoptic}, a realistic dataset with diverse and fine-grained STSG annotations, 20BN  \citep{goyal201720bn},  a video dataset focusing on daily actions, and MUGEN \citep{hayes2022mugen}, a synthetic dataset containing gameplay footages. 
These datasets vary significantly in their temporal patterns, showcasing the versatility of STSL.
Specifically, OpenPVSG captions focus on natural and complex events, 20BN captions are in the form of action pre-conditions and post-conditions, while MUGEN captions describe consecutive action sequences performed by the main protagonist in the game.

% Our approach demonstrates substantial improvements over \textit{fully-supervised} baselines. On OpenPVSG, \ours~achieves a unary predicate prediction accuracy of $27.78\%~(+12.65\%)$ and a binary recall@5 of $0.42~(+0.22)$.  
% Furthermore, \ours\ exceeds baselines by $7\%$ on 20BN and $5.2\%$ on MUGEN in terms of overall predicate prediction accuracy.

Our main result shows significant improvements of \ours~over fully supervised baselines. 
On OpenPVSG, \ours\ achieves a unary predicate prediction accuracy of 27.78\% and a binary recall@5 of 0.42, surpassing the best fully supervised baseline by 12.65\% and 0.22, respectively. 
Furthermore, \ours~outperforms baselines by 7\% on 20BN and 5.2\% on MUGEN in overall predicate prediction accuracy. 
We now delve into the experiments conducted on each of these datasets.

% Further experimental details are provided in \appref{app:experimental_details}.

% \input{figures/OpenPVSG_ablation}

\subsection{OpenPVSG Dataset}

\begin{table}[tb]
    \centering
    \small{
\begin{tabular}{c|l||c|c|c||c|c|c}
\toprule
\hline
\multicolumn{2}{l||}{\multirow{2}{*}{\textbf{Method} (with LASER)}} & \multicolumn{3}{c||}{\textbf{ Unary}} & \multicolumn{3}{c}{\textbf{Binary}} \\ \cline{3-8} 
\multicolumn{1}{l}{} & \multicolumn{1}{l||}{} & \textbf{ R@1} & \textbf{R@5} & \textbf{R@10} & \textbf{R@1} & \textbf{R@5} & \textbf{R@10} \\ \hline
\multirow{3}{*}{VIOLET}
& Base & 0.0660 & 0.1855 & 0.2983 & 0.0460 & 0.1307 & 0.2636\\
& Fine-tuned &0.0878 & 0.2574 & 0.3463 & 0.0501 & 0.2028 & 0.3451\\
& Incr. & \color{sclgreen} {$\uparrow$ 0.0218} & \color{sclgreen} {$\uparrow$ 0.0719} & \color{sclgreen} {$\uparrow$ 0.0480} & \color{sclgreen} {$\uparrow$ 0.0041 } & \color{sclgreen} {$\uparrow$ 0.0721} & \color{sclgreen} {$\uparrow$ 0.0815}\\ \hline
\multirow{3}{*}{SigLIP}
& Base & 0.0000 & 0.0179 & 0.0483 & 0.0000 & 0.0362 & 0.1667\\
& Fine-tuned & 0.1467 & 0.2627 & 0.3152 & 0.0347 & 0.1624 & 0.3012\\
& Incr. & {\color{sclgreen}{$\uparrow$ 0.1467}} & \color{sclgreen} {$\uparrow$ 0.2448} & \color{sclgreen} {$\uparrow$ 0.2669} & \color{sclgreen} {$\uparrow$ 0.0347} & \color{sclgreen} {$\uparrow$ 0.1262} & \color{sclgreen} {$\uparrow$ 0.1345}\\ \hline
\multirow{3}{*}{CLIP}
& Base &0.1633 & 0.3381 & 0.4404 & 0.0197 & 0.0673 & 0.0988\\
& Fine-tuned &0.2778 & 0.5231 & 0.6402 & 0.1482 & 0.4214 & 0.5398\\ 
& Incr. & \color{sclgreen} {$\uparrow$ 0.1145} & \color{sclgreen} {$\uparrow$ 0.1850} & \color{sclgreen} {$\uparrow$ 0.1998} & \color{sclgreen} {$\uparrow$ 0.1284 } & \color{sclgreen} {$\uparrow$ 0.3540} & \color{sclgreen} {$\uparrow$ 0.4410}\\ \hline
\bottomrule                
\end{tabular}
}
\caption{
We show the performance improvements of base backbone models and their fine-tuned version, on the R@k metrics of unary and binary predicate prediction.
As shown by the increments, \ours's weak supervisory learning framework significantly enhances all three models' performance on the STSG extraction tasks.
% \todo{Do insight and qualitative studies.}
}
\label{tab:openpvsg_backbone}
\vspace{-10px}
\end{table}

\textbf{Dataset.}
The OpenPVSG \citep{yang2023panoptic} dataset comprises $400$ videos sourced from Ego4D \citep{grauman2021ego4d}, VidOr \citep{shang2019annotating, thomee2016yfcc100m}, and EpicKitchen \citep{Damen2022RESCALING, Damen2018EPICKITCHENS}. 
This dataset offers fine-grained ground truth annotations of STSGs for $150K$ frames, encompassing $126$ object classes and $57$ relation classes.
% Each video is accompanied by multiple captions with their corresponding video snippets. 
We train on $1,832$ video-caption pairs, and evaluate on $438$ video-STSG pairs.

% \textbf{Temporal Specifications.}
% We obtain the temporal specifications for the OpenPVSG dataset through  3-shot GPT-4. 

% GPT-4 takes in a caption and generates a series of events in json format, and we postprocess the events into programmatic spatio-temporal specifications in STSL.

\textbf{Experimental Setup.}
As illustrated in Figure \ref{fig:model-architecture}, our objective is to train an STSG generator capable of taking a video clip with object bounding boxes as input and predicting the properties, attributes, and relationships between objects. 
We leverage LASER to fine-tune three vision-language models—VIOLET \citep{fu2021violet}, SigLIP  \citep{zhai2023sigmoid}, and CLIP  \citep{radford2021clip}—using weak supervision from captions.
These models predict both similarity scores between cropped objects and unary predicate keywords, as well as between object pairs and binary predicate keywords, resulting in a probabilistic STSG. 
All backbone models support open-world vocabularies and are thus robust to the fuzziness present in GPT-4-generated structured representations.

\begin{figure}[t!]
\begin{minipage}{0.42\textwidth}
    \centering
    \includegraphics[width=\linewidth]{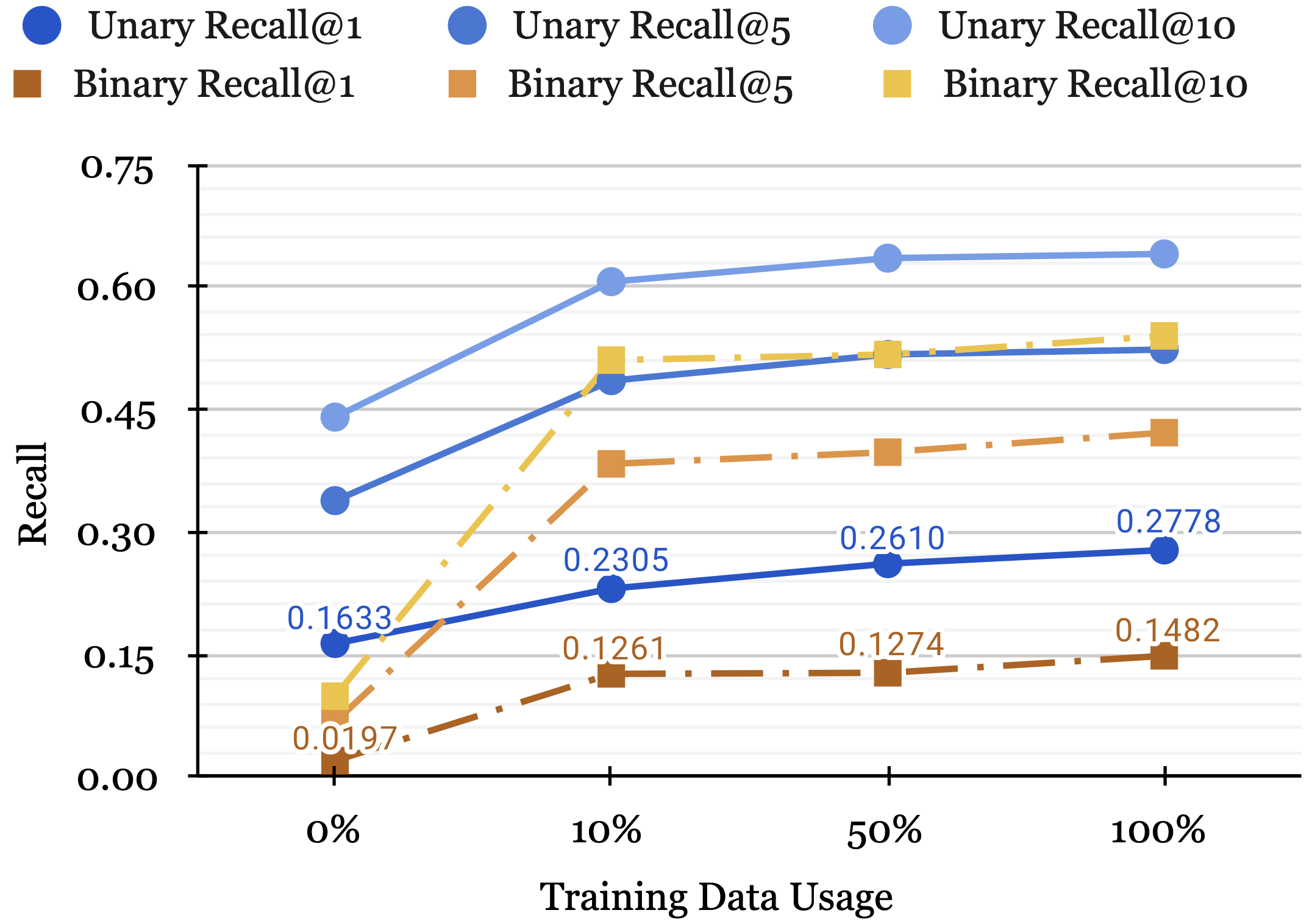}
    \captionof{figure}{Data-efficient fine-tuning on OpenPVSG dataset with LASER: Providing only 10\%, 50\%, and 100\% of the training dataset significantly enhances the performance of CLIP model. }
    \label{fig:openpvsg_data_efficient}
\end{minipage}
\hfill
\begin{minipage}{0.56\textwidth}
\footnotesize
\begin{tabular}{l||c||c|c|c}
\toprule
\hline
 \multicolumn{1}{l||}{\multirow{2}{*}{\textbf{Method}}} & \multicolumn{1}{c||}{\textbf{Unary}}          & \multicolumn{3}{c}{\textbf{Binary}}\\ \cline{2-5} 
 & \textbf{Acc.} ($\%$) & \textbf{R@1} & \textbf{ R@5} & \textbf{ R@10} \\ \hline
 IPS-Vanilla & 15.13 & 0.0741 & 0.1081 & 0.1109\\
 IPS-Filter & 13.14  & 0.0777 & 0.1040 & 0.1133\\
 IPS-Conv & 15.13  & 0.0861 & 0.1143 & 0.1218\\
 IPS-Trans & 14.67  &0.1419 & 0.2032 & 0.2207\\
 VPS-Vanilla & 5.49  & 0.0374 & 0.0517 & 0.0531\\
 VPS-Filter & 5.46 & 0.0405 & 0.0480 & 0.0488\\
 VPS-Conv & 7.46 & \textbf{0.1616} & 0.1781 & 0.2343\\
 VPS-Trans & 5.46  & 0.1019 & 0.1499 & 0.1562\\
 \hline
 LASER-CLIP & \textbf{27.78} & 0.1482 & \textbf{0.4214} & \textbf{0.5398} \\ \hline
\bottomrule                
\end{tabular}
\captionof{table}{
Comparison between \textit{weakly supervised} LASER-CLIP and \textit{fully supervised} IPS and VPS methods on various backbones trained on the full OpenPVSG. 
LASER-CLIP significantly outperforms all baselines, except on Binary R@1, despite using \textit{weak supervision}.
}
\label{tab:openpvsg_supervised}
\end{minipage}
% \vspace{-15px}
\end{figure}

\textbf{Evaluation Metric.}
We evaluate model performance using Recall@$k$ (R@$k$) which estimates whether the ground truth label is within the top-$k$ prediction of a given model. 
During evaluation, the model processes (a) the full vocabulary of object and relation classes and (b) preprocessed cropped objects and object pairs, predicting the probabilistic STSG.
In particular, unary R@$k$ assesses object category prediction capability, while binary R@$k$ evaluates pair-wise prediction of binary relations. 
% For each object in a video, the model predicts a distribution of categorical labels, and for each pair of related objects in each frame, it predicts a distribution of relation labels. 
% Recall@k measures whether the ground truth label is within the top-k predictions, and overall recall is calculated by comparing these predictions against the annotated ground truth video STSG. 
% Unlike the original VIOU metric from the OpenPVSG, which combines object segmentation, object classification, and relational classification into a single score, R@$k$ provides a more detailed assessment of a model’s ability to recognize objects and relations separately.

\textbf{Backbone models significantly improve after been weakly supervised by LASER.}
% We compare the performance improvements of the back-bone models after fine-tuning to validate the effectiveness of \ours~in learning STSGs with weak supervision labels.
We validate \ours' effectiveness in learning STSGs with weak supervision by comparing the performance improvements of the backbone models after fine-tuning. 
% As shown in \tabref{tab:openpvsg_backbone}, after training on the full OpenPVSG dataset, \ours~significantly improves the performance of the base backbones models using only captions for weak supervision.
As shown in \tabref{tab:openpvsg_backbone}, \ours~significantly enhances backbone performance using only captions for weak supervision on the OpenPVSG dataset.

\textbf{Data efficiency of LASER.}
To further assess \ours's data efficiency, we train the model on 10\% and 50\% of the training dataset. 
As illustrated in \figref{fig:openpvsg_data_efficient}, even with just 10\% of the training data (183 video-caption pairs), \ours~significantly enhances the unary R@1 from 0.1633 to 0.2305 and the binary R@1 from 0.0197 to 0.1261.
On average, using just 10\% of the data achieves 70.75\% of the performance obtained with the full dataset, highlighting LASER's data efficiency.
% We further note that, fine-tuning on 10\% of the data only takes an hour under our server setup, which demonstrates our learning efficiency.

\textbf{Weak-supervision may outperform full-supervision.}
To better understand the efficacy of weak supervision, we also compare them against fully supervised methods. 
We study $8$ fully supervised baselines, which employ two different video panoptic segmentation strategies: Image Panoptic Segmentation with Tracker (IPS) and Video Panoptic Segmentation (VPS). 
For relation extraction, the baselines employ 4 model architectures: (1) Vanilla: fully-connected layers, (2) Filter: handcrafted filters, (3) Conv: 1D-convolutional layers, and (4) Trans: transformer encoders.
As shown in \tabref{tab:openpvsg_supervised}, \ours\ significantly outperforms the best fully supervised methods in all metrics except for binary R@1, where it ranks just after the top-performing VPS-Conv.

\vspace{-0.05in}
\subsection{20BN Dataset}
\begin{figure*}[t]
\centering
\begin{tikzpicture}
\pgfplotsset{compat=1.8}
\pgfplotsset{every tick label/.append style={font=\scriptsize\sf}}
\begin{axis}[
    height=4.5cm,
    width=\textwidth,
    bar width = 0.05cm,
	major x tick style = transparent,
	symbolic x coords={is-bendable, is-fluid, is-holdable, is-rigid, is-tearable, above, attached, behind, broken, close, closed, deformed, empty, far, fits, folded, has-hole, high, in, in-front, left, low, next-to, on, on-surface, open, right, stacked, stretched, torn, twisted, under, upright, visible},
    ymin=0,
    ymax=1.05,
    ytick={0.0,0.20,0.40,...,1.0},
	ylabel=\scriptsize{F1 Score},
	legend style={
                at={(1.0,0.0)},
                anchor=south east,
                legend columns=-1,
                font=\scriptsize,
                rotate=90
        },
	ymajorgrids = true,
	legend cell align=left,
	enlarge x limits=0.03,
	xtick = data,
	ybar,
	y label style={at={(axis description cs:-0.035,.5)}, font={\scriptsize}},
    x label style={
        at={(axis description cs:.5,0.02)}, font=\scriptsize,
    },
    x tick label style={
        anchor=east,
        yshift=1.5mm, 
        rotate=45
    },
]

\addplot+[mydeeporange, fill=myorange]
coordinates{
(is-bendable, 0.89)
(is-fluid, 0.28)
(is-holdable, 1)
(is-rigid, 0.89)
(is-tearable, 0.86)
(above, 0.68)
(attached, 0.16)
(behind, 0.6)
(broken, 0.53)
(close, 0.91)
(closed, 0.71)
(deformed, 0.34)
(empty, 0.56)
(far, 0.16)
(fits, 0.97)
(folded, 0.48)
(has-hole, 0.55)
(high, 0.5)
(in, 0.89)
(in-front, 0.58)
(left, 0.65)
(low, 0.64)
(next-to, 0.64)
(on, 0.62)
(on-surface, 0.9)
(open, 0.58)
(right, 0.58)
(stacked, 0.8)
(stretched, 0.37)
(torn, 0.57)
(twisted, 0.46)
(under, 0.69)
(upright, 0.71)
(visible, 1)};
	
\addplot+[myblue, fill=mylightblue]
coordinates{
(is-bendable, 0.8309455587)
(is-fluid, 0.8677248677)
(is-holdable, 0.9992727273)
(is-rigid, 0.8657648283)
(is-tearable, 0.85091743119266)
(above, 0.7911646586)
(attached, 0.5263803681)
(behind, 0.6884899683)
(broken, 0.3673139159)
(close, 0.9263636364)
(closed, 1.00)
(deformed,0.2479338843)
(empty, 0.9230769231)
(far, 0.1205615194)
(fits, 0.9873228934)
(folded, 0.6656976744)
(has-hole,0.6666666667)
(high, 0.6185804962)
(in, 0.9952272256)
(in-front, 0.6948356808)
(left, 0.9228871067)
(low, 0.6679360244)
(next-to, 0.7749440716)
(on, 0.9971883786)
(on-surface, 0.858983891)
(open, 0.9975429975)
(right, 0.9302325581)
(stacked, 0.7833333333)
(stretched, 0.3801652893)
(torn, 0.6648599819)
(twisted, 0.6402116402)
(under, 0.2259414226)
(upright, 0.4769585253)
(visible, 0.9929742389)};

\addplot+[mydeepblue, fill=myblue]
coordinates{
(above, 0.84)
(attached, 0.7083333333)
(behind, 0.7692307692)
(broken, 0.2105263158)
(close, 0.9322033898)
(closed, 0.8965517241)
(deformed, 0)
(empty, 0.88)
(far, 0.1818181818)
(fits, 1)
(folded, 0.7)
(has-hole, 0)
(high, 0.4255319149)
(in, 1)
(in-front, 0.6666666667)
(is-bendable, 0.7368421053)
(is-fluid, 0.4444444444)
(is-holdable, 1)
(is-rigid, 0.4975609756)
(is-tearable, 0.9019607843)
(left, 0.8041237113)
(low, 0.6037735849)
(next-to, 0.8125)
(on, 1)
(on-surface, 0.8504801097)
(open, 1)
(right, 0.8695652174)
(stacked, 0.8571428571)
(stretched, 0)
(torn, 0.07692307692)
(twisted, 0)
(under, 0.5555555556)
(upright, 0.5081967213)
(visible, 0.9753694581)};

\legend{GPA, \ours-P, \ours}
\end{axis}
\end{tikzpicture}
\vspace{-10px}
\caption{
Per-predicate F1 score performance comparison of LASER, LASER-P, and GPA, all trained on the full 20BN dataset. LASER-P outperforms GPA on 71\% of predicates, and LASER outperforms GPA on 59\% of the predicates. 
}
\label{fig:20bn-f1}
\end{figure*}

\begin{figure*}[tb]
    \centering
    \includegraphics[width=\linewidth]{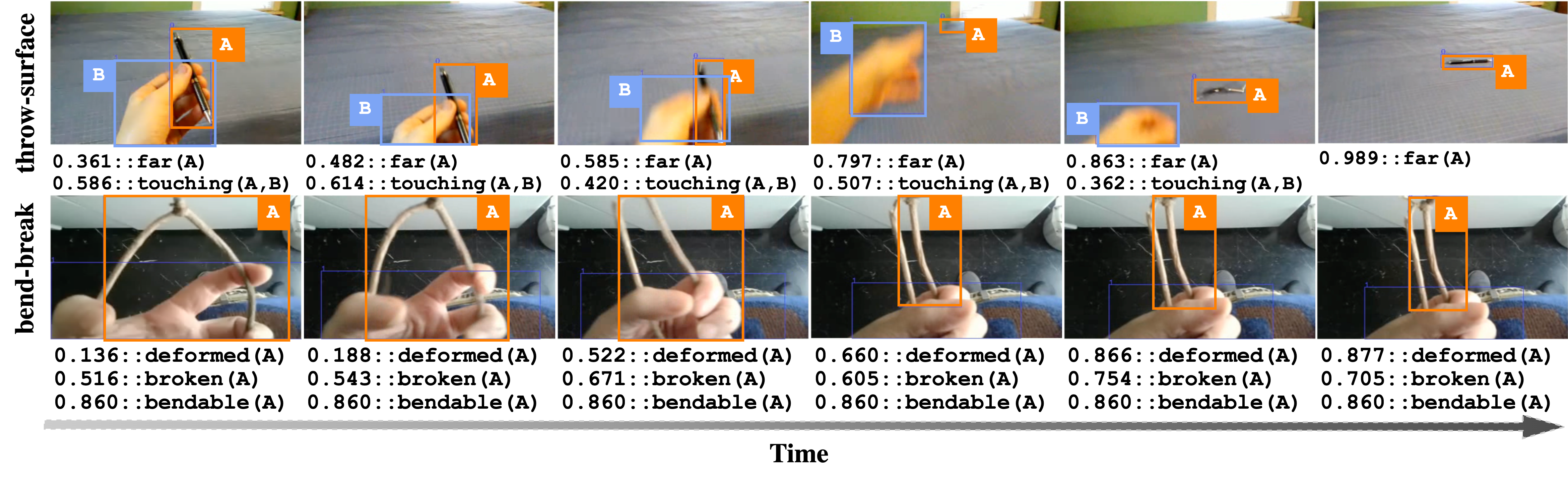}
    \vspace{-10px}
    \caption{
    Qualitative study of the model trained with LASER on the full 20BN dataset. Each row displays a sequence of frames from a video, with bounding boxes labeled by object IDs. The left side of each row shows the action label, while the bottom of each row lists the attributes and relationships associated with the objects, along with the corresponding likelihoods of these facts holding true.
    }
    \label{fig:20bn_qualitative}
    \vspace{-15px}
\end{figure*}

\textbf{Dataset.}
The 20BN dataset consists of 
(a) video and action pairs of humans performing everyday actions with ordinary objects  \citep{goyal201720bn},  
(b) expert designed pre-conditions and post-conditions for the actions in the PDDL language  \citep{migimatsu2022grounding},
and 
(c) frame-based object bounding boxes  \citep{materzynska2020something}.
There are 172 actions with 37 underlying predicates in this dataset, capturing object attributes, object states, and relationships between two objects.
Specifically, there are 6 static predicates, 21 unary predicates, and 10 binary predicates.
We train on $10,000$ training datapoints and test on $14,816$ data points. 
% The full experimental details are presented in \appref{app:20bn_details}.
% where the training and testing splits follow the 20BN dataset.

\textbf{Experimental Setup.}
In the 20BN dataset, each video is assigned an action label from a set of 172 possible actions. 
Each action is annotated with a natural language description and a logical specification, represented as a pair of pre-condition and post-condition in PDDL format. 
These PDDL specifications naturally map to the $\lozenge (\psi_{\text{pre}} \wedge \lozenge \psi_{\text{post}})$ structure in STSL. 
This setup enables us to assess the performance difference between caption-generated specifications and ground truth programmatic specifications. 
\ours~received the generated specifications from natural language captions using GPT-4 as weak supervision label. 
For ablation study, we refer to a variant of our method that uses ground truth specifications as weak supervision labels as \ours-P.

\textbf{Evaluation Metrics.}
We consider exact accuracy and F1-score as the metrics to evaluate STSG generators for 20BN dataset.
Specifically, we compute F1-score and accuracy for each predicate in the vocabulary, and evaluate the weighted average of F1-score and accuracy for overall evaluation.

\textbf{Model Architectures.}
Our backbone STSG generator model comprises an S3D   \citep{xie2018s3d} video encoding model pretrained on Kinetics 400 \citep{kay2017kinetics}, followed by ROIpooler for extract object embeddings, then passed to MLP layers for relation classification.
% The full experimental details including model architecture are presented in \appref{app:20bn_details}.

% Our training methodology incorporates three types of loss functions: contrastive loss, temporal loss, and semantic loss. 
% For semantic loss, we extract commonsense integrity constraints, such as ``a rigid object is not fluid.''
% To evaluate the performance of our STSG extraction model, we use the F1-score as a metric, comparing its predictions against the fine-grained STSG labels derived using heuristics, assuming the pre-condition holds at the beginning and the post-condition holds at the end. 
% These labels are established based on the assumption that the pre-condition holds at the beginning and the post-condition holds at the end, serving as the symbolic ground truth representation for the entire video.

\textbf{Quantitative and Qualitative Study.}
We compare our approach against GPA   \citep{migimatsu2022grounding}, a weakly supervised baseline which uses the ground truth specification as learning signals. 
Our evaluation shows that \ours-P, with the same programmatic supervision, achieves a higher average F1-score of $0.77$ and accuracy of $91\%$, compared to the baseline's $0.74$ F1-score and $76\%$ accuracy. \ours, using only natural language descriptions, achieves a comparable F1-score of $0.73$ and better accuracy of $83\%$.
Furthermore, \ours-P outperforms the baseline in $71\%$ of the fine-grained predicate recognition tasks, and \ours~outperforms $59\%$, as shown in Figure \ref{fig:20bn-f1}.
For the qualitative study, we present the STSG generator's frame-wise predictions on several test data points in \figref{fig:20bn_qualitative}. 
In addition to this, we conduct an ablation study, detailed in \tabref{tab:20bn-ablation-loss}, to evaluate the impact of different components of our loss function design.

\vspace{-0.05in}
\subsection{MUGEN Dataset}

\textbf{Dataset.}
MUGEN \citep{hayes2022mugen} is a synthetic dataset that is based on an open-sourced platform game CoinRun \citep{cobbe2019coinrun}.
Each datapoint contains a $3.2$s video snippet of the game-play and a corresponding automatically generated text describing the video.
The agent in the game can perform $6$ actions: \texttt{walk}, \texttt{jump}, \texttt{kill}, \texttt{collect}, \texttt{die}, and \texttt{climb}.
% There are $13$ different characters and entities in the game, including various enemies and collectibles.
A video may contain any sequence of actions, which may be simple or complex.
We train on $5,000$ training datapoints and test on $12,851$ datapoints.
% Further details are provided in \appref{app:mugen_experiment}.
% , where the training and testing splits are already provided in the dataset.

% \textbf{Temporal Specifications.}
% We construct the temporal specification from the given text description using a simple heuristic-based semantic parser. 
% We first extract an ordered action list $[a_1, a_2, \dots, a_n]$ from the text with heuristics, then construct the temporal specification as $a_1  \mathbf{U} a_2  \dots \mathbf{U} a_n$.
% This specification means the actions are performed one by one with no intermediate gaps.
% The generated specifications are used as weak supervision labels to learn the STSG model.

\textbf{Experimental Setup.}
In the MUGEN dataset, each video is paired with a natural language description that can be represented as an ordered list of actions $[a_1, a_2, \dots, a_n]$. 
The corresponding STSL specifications are naturally expressed in the logical form $a_1 \mathbf{U} a_2 \dots \mathbf{U} a_n$. 
\ours~uses these generated specifications, derived from natural language captions via a crafted semantic parser, as weak supervision labels. 
We evaluate the model's performance based on action prediction accuracy.

\textbf{Model Architectures.}
Since the MUGEN videos differ significantly from natural videos, our backbone model is trained from scratch. 
Our backbone STSG generator model comprises an S3D video encoding model pretrained on Kinetics 400, followed by ROIpooler for extract object embeddings, then passed to MLP layers for relation classification.

\textbf{Quantitative Study.}
We evaluate our action prediction performance on MUGEN and compare it to caption supervised VT-TWINS  \citep{ko2022video}, TempCLR \citep{yang2023tempclr}, and a directly supervised baseline using the same model architecture as ours but with ground truth labels (Supervised). 
We provide all baselines with extra annotations on the start- and end-frame of each clip-based textual description.
% As the ground truth actions for the video are different from the text description, we use heuristics to obtain the start and end frames.
% For a fair comparison, we fine-tune an S3D backbone pre-trained using Kinetics 400 (the same as \ours) along with the baseline models.
% This label is used for training both VT-TWINS and fully supervised models.
As shown in Table \ref{tab:performance}, \ours~has better action prediction accuracy despite receiving less supervision than the baselines.
Moreover, our weakly supervised model even achieves better accuracy than the Supervised method on when trained on $5000$ data points.
% We further evaluate our approach on a downstream video-specification retrieval task. 
% Given 3 videos and 3 specifications, we want to match the correctly aligned pairs, retrieve specification given video and vice versa.
% We denote the two tasks as spec-retrieval and video-retrieval respectively.
% Specifically, we evaluate the accuracy with which we infer whether the correct specification has the highest alignment score, and vice versa.
% \ours{} outperforms an embedding-based baseline SDSC \citep{hayes2022mugen} on both tasks (Table~\ref{tab:downstream}).
% Our approach can even identify actions that persist for a very short period of time, such as \texttt{kill} an enemy.

% As shown in \tabref{tab:quantitative}, \ours\ outperforms the SDSC baseline \citep{hayes2022mugen}, an embedding-based approach presented along with the MUGEN dataset. 
% SDSC learns a projection from video and text embeddings to a joint embedding space and uses a scaled cosine distance function to measure the similarity.
% SDSC utilizes the same pre-trained video embedding model S3D and DistilBERT\citep{sanh2019distilbert} for specification embeddings.
% Compared to the baseline, \ours\ achieves better performance on both specification and video retrieval tasks.

% \textbf{Qualitative Study.}
% \ours\ provides clip-level explanations along with the retrieval results. 
% We illustrate the fine-grained model predictions on two test datapoints in \figref{fig:mugen_qualitative}.
% Our approach can even identify actions that persist for a very short period of time, such as \texttt{kill} an enemy.

\begin{table}[t]
% \footnotesize
    \scriptsize
    \begin{minipage}{\linewidth}
        \begin{minipage}{.43\linewidth}
            \begin{center}
            \begin{tabular}{ c||c|c|c||c  } 
            \toprule
            \hline
            \textbf{Setup} &
            \textbf{Contr.} &
            \textbf{Sem.} &
            \textbf{Tmp.} &
            \textbf{F1} \\
            \hline 
            1 & \checkmark &  &  & 0.31  \\
            2 & \checkmark & \checkmark &  & 0.31  \\
            3 & \checkmark & & \checkmark & 0.58 \\ \midrule
            Full & \checkmark & \checkmark & \checkmark & \textbf{0.77} \\
            \hline 
            \bottomrule
            \end{tabular}
            \end{center}
            \vspace{-10px}
            \caption{
            Ablation study on loss function components (\underline{contr}astive, \underline{sem}antic, and \underline{t}e\underline{mp}oral losses), trained and evaluated on the full 20BN dataset.
            By ablating on loss function components, we find all are essential to the performance of our method.
            }
            \label{tab:20bn-ablation-loss}
        \end{minipage}
        \hfill
        \begin{minipage}{.55\linewidth}
            \begin{minipage}{\linewidth}
                \centering
                \begin{tabular}{ c||c|c|c|| c } 
                    \toprule
                    \hline
                    \textbf{\#Data} & \textbf{\ours} & \textbf{VT-TWINS} & \textbf{TempCLR} & \textbf{Supervised} \\
                    \hline
                    % 100 & $\mathbf{46.0}$ & $37.0$& $16.4$ &$52.0$\\
                    % 1000 & $\mathbf{48.5}$ & $22.8$ & $17.10$ & $52.0$  \\
                    % 5000 & $\mathbf{59.6}$  & $55.6$ & $34.91$ & $54.4$\\
                    % 100 & $\mathbf{46.0} \pm 9.7$ & $37.0 \pm 26.3$ & $16.4 \pm 26.7$ &$52.0 \pm 0.8$\\
                    % 1000 & $\mathbf{48.5} \pm 2.0$ & $22.8 \pm 9.6$ & $17.10 \pm 18.41 $ & $52.0 \pm 0.7$  \\
                    % 5000 & $\mathbf{59.6} \pm 9.7$  & $55.6 \pm 13.7$ & $34.91 \pm 9.41 $ & $54.4 \pm 0.1$\\
                    100 & $\mathbf{46.0}$ & $37.0$ & $16.4$ &$52.0$\\
                    1000 & $\mathbf{48.5}$ & $22.8$ & $17.10$ & $52.0$  \\
                    5000 & $\mathbf{59.6}$  & $55.6$ & $34.91$ & $54.4$\\
                    \hline
                    \bottomrule
                \end{tabular}
                \vspace{8px}
                \caption{
                Data efficiency study of the LASER model trained on the MUGEN dataset, compared against VT-twins, TEMP-CLR, and fully supervised baselines. The models were trained on 100, 1,000, and 5,000 data points from the MUGEN dataset, evaluated using action prediction accuracy (\%).
                }
                \label{tab:performance}
            \end{minipage}
            \hspace{1cm}
            % \begin{minipage}{\linewidth}
            % \centering
            %     \begin{tabular}{ c||c|c  } 
            %     \toprule
            %     \hline
            %     \textbf{Task} & \textbf{SDSC} & \oursbf-\textbf{P} \\
            %     \hline
            %     Spec-Ret. & 87.00\% & \textbf{93.80\%} \\
            %     Vid-Ret. & 86.80\% & \textbf{90.00\%} \\
            %     \bottomrule
            %     \end{tabular}
            %     \caption{
            %     Comparison to baselines on downstream tasks of video- or spec-retrieval with the MUGEN dataset.
            %     }
            %     \label{tab:downstream}
            % \end{minipage}
        \end{minipage}
    \end{minipage}
\vspace{-10px}
\end{table}

% \input{figures/mugen_downstream}
% \textbf{Application to Downstream Tasks.}
% To further illustrate the usefulness of the learnt video representation, we evaluate our approach on a downstream video-specification retrieval task. 
% Given three videos and three specifications, we want to match the correctly aligned pairs.
% We denote the two tasks as spec-retrieval and video-retrieval respectively.
% Specifically, we evaluate the accuracy with which we infer whether the correct specification has the highest alignment score, and vice versa.
% \ours~ outperforms an embedding-based baseline SDSC \citep{hayes2022mugen} on both tasks (Table \ref{tab:downstream}).

% \input{sections/6_related_work}
\section{Conclusion, Limitation, and Future Outlook}

% We present \ours, which comprises a generic specification language STSL and a differentiable neuro-symbolic framework for learning semantic video representations.
% Our approach captures the rich spatial and temporal properties of a video by learning a spatio-temporal scene graph (STSG) through weak supervision in the form of logic specifications.

We propose \ours, a model-agnostic, end-to-end differentiable framework for learning spatio-temporal scene graphs with weak supervision from video captions.
We evaluate our work on three datasets OpenPVSG, 20BN, and MUGEN  which exhibit diverse temporal properties. 
\ours\ significantly outperforms even the fully supervised baselines on STSG generation tasks.

\textbf{Limitations.} 
\ours\ still faces scalability challenges regarding both video duration and the number of objects present in the video. 
Learning STSGs over long time horizons with weak signals remains an open problem. 
Additionally, \ours's performance is limited by the quality of the video captions. 
It is unclear whether more specifications can be elicited from LLMs to reduce reliance solely on captions. 
We aim to explore these directions in future work.

% We design STSL, a general and expressive spatio-temporal specification language, for specifying fine-grained video semantics.
% We implement a differentiable neuro-symbolic alignment checker to relate between an STSL specification and a spatio-temporal scene graph.
    
% In the future, we plan to train on a large number of diverse video-caption pairs to develop a foundation model for STSG extraction.

% In the future, we plan to close the gap between natural language and fine-grained logic specification by fine-tuning large language models and incorporating common-sense knowledge bases.
% Further, we want to encourage and inspire the neural symbolic community to push the boundary of video representation learning.

% The current limitations of \ours\ include the requirement for pre-specifying the relational schema and providing complex temporal specifications.
% In the future, we plan to extend our approach by incorporating the natural language modality into our end-to-end pipeline, i.e., learning to systematically extract the schema and the spatio-temporal specifications from natural language captions and common-sense knowledge bases.

% \todo{ First paper}
% \todo{Future work: Caption LLM, fine grained caption}
% \todo{Future work: Auto formalization to specs}
\section*{Acknowledgements}

We sincerely thank the anonymous reviewers for their valuable feedback and constructive suggestions. 
We also gratefully acknowledge Mayank Keoliya, Matthew Kuo, Amish Sethi, and Neelay Velingker for their help on the project.
This research was supported by ARPA-H program on Safe and Explainable AI under the award D24AC00253-00 and NSF award CCF 2313010.

% \newpage
\small
\bibliographystyle{iclr2025_conference}
\bibliography{iclr2025_main}

\begin{thebibliography}{60}
\providecommand{\natexlab}[1]{#1}
\providecommand{\url}[1]{\texttt{#1}}
\expandafter\ifx\csname urlstyle\endcsname\relax
  \providecommand{\doi}[1]{doi: #1}\else
  \providecommand{\doi}{doi: \begingroup \urlstyle{rm}\Url}\fi

\bibitem[Albers et~al.(2009)Albers, Marchetti-Spaccamela, Matias, Nikoletseas, and Thomas]{albers2009automata}
Susanne Albers, Alberto Marchetti-Spaccamela, Yossi Matias, Sotiris Nikoletseas, and Wolfgang Thomas.
\newblock \emph{Automata, Languages and Programming: 36th International Colloquium, ICALP 2009, Rhodes, Greece, July 5-12, 2009, Proceedings, Part I}, volume 5555.
\newblock Springer Science \& Business Media, 2009.

\bibitem[Apriceno et~al.(2022)Apriceno, Passerini, and Serafini]{apriceno2022neuro}
Gianluca Apriceno, Andrea Passerini, and Luciano Serafini.
\newblock A neuro-symbolic approach for real-world event recognition from weak supervision.
\newblock In \emph{29th International Symposium on Temporal Representation and Reasoning}. Schloss Dagstuhl-Leibniz-Zentrum f{\"u}r Informatik, 2022.

\bibitem[Chaki et~al.(2005)Chaki, Clarke, Ouaknine, Sharygina, and Sinha]{chaki2005concurrent}
Sagar Chaki, Edmund Clarke, Jo{\"e}l Ouaknine, Natasha Sharygina, and Nishant Sinha.
\newblock Concurrent software verification with states, events, and deadlocks.
\newblock \emph{Formal Aspects of Computing}, 17\penalty0 (4):\penalty0 461--483, 2005.

\bibitem[Chang et~al.(2019)Chang, Huang, Sui, Fei-Fei, and Niebles]{chang2019d3tw}
Chien-Yi Chang, De-An Huang, Yanan Sui, Li~Fei-Fei, and Juan~Carlos Niebles.
\newblock D3tw: Discriminative differentiable dynamic time warping for weakly supervised action alignment and segmentation.
\newblock In \emph{Proceedings of the IEEE/CVF Conference on Computer Vision and Pattern Recognition}, pp.\  3546--3555, 2019.

\bibitem[Cobbe et~al.(2019)Cobbe, Klimov, Hesse, Kim, and Schulman]{cobbe2019coinrun}
Karl Cobbe, Oleg Klimov, Chris Hesse, Taehoon Kim, and John Schulman.
\newblock Quantifying generalization in reinforcement learning.
\newblock In \emph{International Conference on Machine Learning}, pp.\  1282--1289. PMLR, 2019.

\bibitem[Cong et~al.(2021)Cong, Liao, Ackermann, Yang, and Rosenhahn]{Cong2021sttran}
Yuren Cong, Wentong Liao, Hanno Ackermann, Michael~Ying Yang, and Bodo Rosenhahn.
\newblock Spatial-temporal transformer for dynamic scene graph generation.
\newblock \emph{CoRR}, abs/2107.12309, 2021.

\bibitem[Damen et~al.(2018)Damen, Doughty, Farinella, Fidler, Furnari, Kazakos, Moltisanti, Munro, Perrett, Price, and Wray]{Damen2018EPICKITCHENS}
Dima Damen, Hazel Doughty, Giovanni~Maria Farinella, Sanja Fidler, Antonino Furnari, Evangelos Kazakos, Davide Moltisanti, Jonathan Munro, Toby Perrett, Will Price, and Michael Wray.
\newblock Scaling egocentric vision: The epic-kitchens dataset.
\newblock In \emph{European Conference on Computer Vision}, 2018.

\bibitem[Damen et~al.(2022)Damen, Doughty, Farinella, Furnari, Ma, Kazakos, Moltisanti, Munro, Perrett, Price, and Wray]{Damen2022RESCALING}
Dima Damen, Hazel Doughty, Giovanni~Maria Farinella, Antonino Furnari, Jian Ma, Evangelos Kazakos, Davide Moltisanti, Jonathan Munro, Toby Perrett, Will Price, and Michael Wray.
\newblock Rescaling egocentric vision: Collection, pipeline and challenges for epic-kitchens-100.
\newblock \emph{International Journal of Computer Vision}, 130:\penalty0 33–55, 2022.

\bibitem[De~Giacomo \& Vardi(2013)De~Giacomo and Vardi]{de2013linear}
Giuseppe De~Giacomo and Moshe~Y Vardi.
\newblock Linear temporal logic and linear dynamic logic on finite traces.
\newblock In \emph{IJCAI'13 Proceedings of the Twenty-Third international joint conference on Artificial Intelligence}, pp.\  854--860. Association for Computing Machinery, 2013.

\bibitem[Ding \& Xu(2018)Ding and Xu]{ding2018weakly}
Li~Ding and Chenliang Xu.
\newblock Weakly-supervised action segmentation with iterative soft boundary assignment.
\newblock In \emph{Proceedings of the IEEE conference on computer vision and pattern recognition}, pp.\  6508--6516, 2018.

\bibitem[Ding et~al.(2014)Ding, Smith, Belta, and Rus]{ding2014optimal}
Xuchu Ding, Stephen~L Smith, Calin Belta, and Daniela Rus.
\newblock Optimal control of markov decision processes with linear temporal logic constraints.
\newblock \emph{IEEE Transactions on Automatic Control}, 59\penalty0 (5):\penalty0 1244--1257, 2014.

\bibitem[Dvornik et~al.(2021)Dvornik, Hadji, Derpanis, Garg, and Jepson]{dvornik2021drop}
Nikita Dvornik, Isma Hadji, Konstantinos~G. Derpanis, Animesh Garg, and Allan~D. Jepson.
\newblock Drop-dtw: Aligning common signal between sequences while dropping outliers.
\newblock \emph{CoRR}, abs/2108.11996, 2021.

\bibitem[Dwibedi et~al.(2019)Dwibedi, Aytar, Tompson, Sermanet, and Zisserman]{dwibedi2019temp}
Debidatta Dwibedi, Yusuf Aytar, Jonathan Tompson, Pierre Sermanet, and Andrew Zisserman.
\newblock Temporal cycle-consistency learning.
\newblock \emph{CoRR}, abs/1904.07846, 2019.

\bibitem[Fu et~al.(2021)Fu, Li, Gan, Lin, Wang, Wang, and Liu]{fu2021violet}
Tsu-Jui Fu, Linjie Li, Zhe Gan, Kevin Lin, William~Yang Wang, Lijuan Wang, and Zicheng Liu.
\newblock Violet: End-to-end video-language transformers with masked visual-token modeling.
\newblock \emph{arXiv preprint arXiv:2111.12681}, 2021.

\bibitem[Goyal et~al.(2017)Goyal, Kahou, Michalski, Materzynska, Westphal, Kim, Haenel, Fr{\"{u}}nd, Yianilos, Mueller{-}Freitag, Hoppe, Thurau, Bax, and Memisevic]{goyal201720bn}
Raghav Goyal, Samira~Ebrahimi Kahou, Vincent Michalski, Joanna Materzynska, Susanne Westphal, Heuna Kim, Valentin Haenel, Ingo Fr{\"{u}}nd, Peter Yianilos, Moritz Mueller{-}Freitag, Florian Hoppe, Christian Thurau, Ingo Bax, and Roland Memisevic.
\newblock The "something something" video database for learning and evaluating visual common sense.
\newblock \emph{CoRR}, abs/1706.04261, 2017.

\bibitem[Grauman et~al.(2021)Grauman, Westbury, Byrne, Chavis, Furnari, Girdhar, Hamburger, Jiang, Liu, Liu, Martin, Nagarajan, Radosavovic, Ramakrishnan, Ryan, Sharma, Wray, Xu, Xu, Zhao, Bansal, Batra, Cartillier, Crane, Do, Doulaty, Erapalli, Feichtenhofer, Fragomeni, Fu, Fuegen, Gebreselasie, Gonz{\'{a}}lez, Hillis, Huang, Huang, Jia, Khoo, Kol{\'{a}}r, Kottur, Kumar, Landini, Li, Li, Li, Mangalam, Modhugu, Munro, Murrell, Nishiyasu, Price, Puentes, Ramazanova, Sari, Somasundaram, Southerland, Sugano, Tao, Vo, Wang, Wu, Yagi, Zhu, Arbel{\'{a}}ez, Crandall, Damen, Farinella, Ghanem, Ithapu, Jawahar, Joo, Kitani, Li, Newcombe, Oliva, Park, Rehg, Sato, Shi, Shou, Torralba, Torresani, Yan, and Malik]{grauman2021ego4d}
Kristen Grauman, Andrew Westbury, Eugene Byrne, Zachary Chavis, Antonino Furnari, Rohit Girdhar, Jackson Hamburger, Hao Jiang, Miao Liu, Xingyu Liu, Miguel Martin, Tushar Nagarajan, Ilija Radosavovic, Santhosh~Kumar Ramakrishnan, Fiona Ryan, Jayant Sharma, Michael Wray, Mengmeng Xu, Eric~Zhongcong Xu, Chen Zhao, Siddhant Bansal, Dhruv Batra, Vincent Cartillier, Sean Crane, Tien Do, Morrie Doulaty, Akshay Erapalli, Christoph Feichtenhofer, Adriano Fragomeni, Qichen Fu, Christian Fuegen, Abrham Gebreselasie, Cristina Gonz{\'{a}}lez, James Hillis, Xuhua Huang, Yifei Huang, Wenqi Jia, Weslie Khoo, J{\'{a}}chym Kol{\'{a}}r, Satwik Kottur, Anurag Kumar, Federico Landini, Chao Li, Yanghao Li, Zhenqiang Li, Karttikeya Mangalam, Raghava Modhugu, Jonathan Munro, Tullie Murrell, Takumi Nishiyasu, Will Price, Paola~Ruiz Puentes, Merey Ramazanova, Leda Sari, Kiran Somasundaram, Audrey Southerland, Yusuke Sugano, Ruijie Tao, Minh Vo, Yuchen Wang, Xindi Wu, Takuma Yagi, Yunyi Zhu, Pablo Arbel{\'{a}}ez, David Crandall, Dima
  Damen, Giovanni~Maria Farinella, Bernard Ghanem, Vamsi~Krishna Ithapu, C.~V. Jawahar, Hanbyul Joo, Kris Kitani, Haizhou Li, Richard~A. Newcombe, Aude Oliva, Hyun~Soo Park, James~M. Rehg, Yoichi Sato, Jianbo Shi, Mike~Zheng Shou, Antonio Torralba, Lorenzo Torresani, Mingfei Yan, and Jitendra Malik.
\newblock Ego4d: Around the world in 3, 000 hours of egocentric video.
\newblock \emph{CoRR}, abs/2110.07058, 2021.

\bibitem[Han et~al.(2022)Han, Xie, and Zisserman]{han2022temporal}
Tengda Han, Weidi Xie, and Andrew Zisserman.
\newblock Temporal alignment networks for long-term video.
\newblock In \emph{Proceedings of the IEEE/CVF Conference on Computer Vision and Pattern Recognition}, pp.\  2906--2916, 2022.

\bibitem[Hayes et~al.(2022)Hayes, Zhang, Yin, Pang, Sheng, Yang, Ge, Hu, and Parikh]{hayes2022mugen}
Thomas Hayes, Songyang Zhang, Xi~Yin, Guan Pang, Sasha Sheng, Harry Yang, Songwei Ge, Qiyuan Hu, and Devi Parikh.
\newblock Mugen: A playground for video-audio-text multimodal understanding and generation.
\newblock \emph{10.48550/ARXIV.2204.08058}, 2022.

\bibitem[Huang et~al.(2020)Huang, Smith, Bastani, Singh, Albarghouthi, and Naik]{huang2020referexpr}
Jiani Huang, Calvin Smith, Osbert Bastani, Rishabh Singh, Aws Albarghouthi, and Mayur Naik.
\newblock Generating programmatic referring expressions via program synthesis.
\newblock In Hal~Daumé III and Aarti Singh (eds.), \emph{Proceedings of the 37th International Conference on Machine Learning}, volume 119 of \emph{Proceedings of Machine Learning Research}, pp.\  4495--4506. PMLR, 13--18 Jul 2020.

\bibitem[Huang et~al.(2021)Huang, Li, Chen, Samel, Naik, Song, and Si]{huang2021scallop}
Jiani Huang, Ziyang Li, Binghong Chen, Karan Samel, Mayur Naik, Le~Song, and Xujie Si.
\newblock Scallop: From probabilistic deductive databases to scalable differentiable reasoning.
\newblock \emph{Advances in Neural Information Processing Systems}, 34:\penalty0 25134--25145, 2021.

\bibitem[Ji et~al.(2021)Ji, Desai, and Niebles]{ji2021detecting}
Jingwei Ji, Rishi Desai, and Juan~Carlos Niebles.
\newblock Detecting human-object relationships in videos.
\newblock In \emph{Proceedings of the IEEE/CVF international conference on computer vision}, pp.\  8106--8116, 2021.

\bibitem[Jia et~al.(2021)Jia, Yang, Xia, Chen, Parekh, Pham, Le, Sung, Li, and Duerig]{jia2021scaling}
Chao Jia, Yinfei Yang, Ye~Xia, Yi{-}Ting Chen, Zarana Parekh, Hieu Pham, Quoc~V. Le, Yun{-}Hsuan Sung, Zhen Li, and Tom Duerig.
\newblock Scaling up visual and vision-language representation learning with noisy text supervision.
\newblock \emph{CoRR}, abs/2102.05918, 2021.

\bibitem[Kay et~al.(2017)Kay, Carreira, Simonyan, Zhang, Hillier, Vijayanarasimhan, Viola, Green, Back, Natsev, et~al.]{kay2017kinetics}
Will Kay, Joao Carreira, Karen Simonyan, Brian Zhang, Chloe Hillier, Sudheendra Vijayanarasimhan, Fabio Viola, Tim Green, Trevor Back, Paul Natsev, et~al.
\newblock The kinetics human action video dataset.
\newblock \emph{arXiv preprint arXiv:1705.06950}, 2017.

\bibitem[Kesten et~al.(1998)Kesten, Pnueli, and Raviv]{kesten1998algorithmic}
Yonit Kesten, Amir Pnueli, and Li-on Raviv.
\newblock Algorithmic verification of linear temporal logic specifications.
\newblock In \emph{Automata, Languages and Programming: 25th International Colloquium, ICALP 1998, Aalborg, Denmark, July 13--17, 1998 Proceedings 25}, pp.\  1--16. Springer, 1998.

\bibitem[Ko et~al.(2022)Ko, Choi, Ko, Noh, On, Kim, and Kim]{ko2022video}
Dohwan Ko, Joonmyung Choi, Juyeon Ko, Shinyeong Noh, Kyoung-Woon On, Eun-Sol Kim, and Hyunwoo~J Kim.
\newblock Video-text representation learning via differentiable weak temporal alignment.
\newblock In \emph{Proceedings of the IEEE/CVF Conference on Computer Vision and Pattern Recognition}, pp.\  5016--5025, 2022.

\bibitem[Kuznetsova et~al.(2018)Kuznetsova, Rom, Alldrin, Uijlings, Krasin, Pont{-}Tuset, Kamali, Popov, Malloci, Duerig, and Ferrari]{kuznetsova2018imagev4}
Alina Kuznetsova, Hassan Rom, Neil Alldrin, Jasper R.~R. Uijlings, Ivan Krasin, Jordi Pont{-}Tuset, Shahab Kamali, Stefan Popov, Matteo Malloci, Tom Duerig, and Vittorio Ferrari.
\newblock The open images dataset {V4:} unified image classification, object detection, and visual relationship detection at scale.
\newblock \emph{CoRR}, abs/1811.00982, 2018.

\bibitem[Lee et~al.(2023)Lee, Sioutis, Ahrens, Alirezaie, Kerzel, and Wermter]{lee2023neuro}
Jae~Hee Lee, Michael Sioutis, Kyra Ahrens, Marjan Alirezaie, Matthias Kerzel, and Stefan Wermter.
\newblock Neuro-symbolic spatio-temporal reasoning.
\newblock In \emph{Compendium of Neurosymbolic Artificial Intelligence}, pp.\  410--429. IOS Press, 2023.

\bibitem[Li et~al.(2021)Li, Selvaraju, Gotmare, Joty, Xiong, and Hoi]{li2021align}
Junnan Li, Ramprasaath~R. Selvaraju, Akhilesh~Deepak Gotmare, Shafiq~R. Joty, Caiming Xiong, and Steven C.~H. Hoi.
\newblock Align before fuse: Vision and language representation learning with momentum distillation.
\newblock \emph{CoRR}, abs/2107.07651, 2021.

\bibitem[Li et~al.(2022)Li, Guo, Liu, and Sun]{li2022essgg}
Xinghang Li, Di~Guo, Huaping Liu, and Fuchun Sun.
\newblock Embodied semantic scene graph generation.
\newblock In Aleksandra Faust, David Hsu, and Gerhard Neumann (eds.), \emph{Proceedings of the 5th Conference on Robot Learning}, volume 164 of \emph{Proceedings of Machine Learning Research}, pp.\  1585--1594. PMLR, 08--11 Nov 2022.

\bibitem[Li et~al.(2023)Li, Huang, and Naik]{li2023scallop}
Ziyang Li, Jiani Huang, and Mayur Naik.
\newblock Scallop: A language for neurosymbolic programming.
\newblock \emph{Proc. ACM Program. Lang.}, 7\penalty0 (PLDI), jun 2023.
\newblock \doi{10.1145/3591280}.

\bibitem[Li et~al.(2024)Li, Huang, Liu, Zhu, Zhao, Dodds, Velingker, Alur, and Naik]{vieira2024li}
Ziyang Li, Jiani Huang, Jason Liu, Felix Zhu, Eric Zhao, William Dodds, Neelay Velingker, Rajeev Alur, and Mayur Naik.
\newblock Relational programming with foundation models.
\newblock AAAI'24/IAAI'24/EAAI'24. AAAI Press, 2024.
\newblock ISBN 978-1-57735-887-9.
\newblock \doi{10.1609/aaai.v38i9.28934}.

\bibitem[Liang et~al.(2024)Liang, Liu, Sun, Xia, and Wang]{liang2024ckt}
Nanhao Liang, Yong Liu, Wenfang Sun, Yingwei Xia, and Fan Wang.
\newblock Ckt-rcm: Clip-based knowledge transfer and relational context mining for unbiased panoptic scene graph generation.
\newblock In \emph{ICASSP 2024-2024 IEEE International Conference on Acoustics, Speech and Signal Processing}, pp.\  3570--3574. IEEE, 2024.

\bibitem[Liu et~al.(2021)Liu, Yan, Mortazavi, and Bhanu]{Liu2021convsgg}
Hengyue Liu, Ning Yan, Masood Mortazavi, and Bir Bhanu.
\newblock Fully convolutional scene graph generation.
\newblock In \emph{Proceedings of the IEEE/CVF Conference on Computer Vision and Pattern Recognition}, pp.\  11546--11556, June 2021.

\bibitem[Lu et~al.(2016)Lu, Krishna, Bernstein, and Fei{-}Fei]{lu2016visual}
Cewu Lu, Ranjay Krishna, Michael~S. Bernstein, and Li~Fei{-}Fei.
\newblock Visual relationship detection with language priors.
\newblock \emph{CoRR}, abs/1608.00187, 2016.

\bibitem[Materzynska et~al.(2020)Materzynska, Xiao, Herzig, Xu, Wang, and Darrell]{materzynska2020something}
Joanna Materzynska, Tete Xiao, Roei Herzig, Huijuan Xu, Xiaolong Wang, and Trevor Darrell.
\newblock Something-else: Compositional action recognition with spatial-temporal interaction networks.
\newblock In \emph{Proceedings of the IEEE/CVF Conference on Computer Vision and Pattern Recognition}, pp.\  1049--1059, 2020.

\bibitem[Miech et~al.(2019)Miech, Alayrac, Smaira, Laptev, Sivic, and Zisserman]{miech2019end}
Antoine Miech, Jean{-}Baptiste Alayrac, Lucas Smaira, Ivan Laptev, Josef Sivic, and Andrew Zisserman.
\newblock End-to-end learning of visual representations from uncurated instructional videos.
\newblock \emph{CoRR}, abs/1912.06430, 2019.

\bibitem[Migimatsu \& Bohg(2022)Migimatsu and Bohg]{migimatsu2022grounding}
Toki Migimatsu and Jeannette Bohg.
\newblock Grounding predicates through actions.
\newblock In \emph{2022 International Conference on Robotics and Automation}, pp.\  3498--3504. IEEE, 2022.

\bibitem[Nag et~al.(2023)Nag, Min, Tripathi, and Chowdhury]{nag2023unbiased}
Sayak Nag, Kyle Min, Subarna Tripathi, and Amit K.~Roy Chowdhury.
\newblock Unbiased scene graph generation in videos.
\newblock \emph{arXiv preprint arXiv:2304.00733}, 2023.

\bibitem[Pnueli(1977)]{Pnueli1977TheTL}
Amir Pnueli.
\newblock The temporal logic of programs.
\newblock \emph{18th Annual Symposium on Foundations of Computer Science}, pp.\  46--57, 1977.

\bibitem[Radford et~al.(2021)Radford, Kim, Hallacy, Ramesh, Goh, Agarwal, Sastry, Askell, Mishkin, Clark, et~al.]{radford2021clip}
Alec Radford, Jong~Wook Kim, Chris Hallacy, Aditya Ramesh, Gabriel Goh, Sandhini Agarwal, Girish Sastry, Amanda Askell, Pamela Mishkin, Jack Clark, et~al.
\newblock Learning transferable visual models from natural language supervision.
\newblock In \emph{International conference on machine learning}, pp.\  8748--8763. PMLR, 2021.

\bibitem[Richard et~al.(2018)Richard, Kuehne, Iqbal, and Gall]{richard2018neuralnetwork}
Alexander Richard, Hilde Kuehne, Ahsan Iqbal, and Juergen Gall.
\newblock Neuralnetwork-viterbi: A framework for weakly supervised video learning.
\newblock In \emph{Proceedings of the IEEE conference on Computer Vision and Pattern Recognition}, pp.\  7386--7395, 2018.

\bibitem[Sadigh et~al.(2014)Sadigh, Kim, Coogan, Sastry, and Seshia]{sadigh2014learning}
Dorsa Sadigh, Eric~S Kim, Samuel Coogan, S~Shankar Sastry, and Sanjit~A Seshia.
\newblock A learning based approach to control synthesis of markov decision processes for linear temporal logic specifications.
\newblock In \emph{53rd IEEE Conference on Decision and Control}, pp.\  1091--1096. IEEE, 2014.

\bibitem[Shang et~al.(2017)Shang, Ren, Guo, Zhang, and Chua]{shang2017video}
Xindi Shang, Tongwei Ren, Jingfan Guo, Hanwang Zhang, and Tat-Seng Chua.
\newblock Video visual relation detection.
\newblock In \emph{Proceedings of the 25th ACM international conference on Multimedia}, pp.\  1300--1308, 2017.

\bibitem[Shang et~al.(2019)Shang, Di, Xiao, Cao, Yang, and Chua]{shang2019annotating}
Xindi Shang, Donglin Di, Junbin Xiao, Yu~Cao, Xun Yang, and Tat-Seng Chua.
\newblock Annotating objects and relations in user-generated videos.
\newblock In \emph{Proceedings of the 2019 on International Conference on Multimedia Retrieval}, pp.\  279--287. ACM, 2019.

\bibitem[Shindo et~al.(2024)Shindo, Brack, Sudhakaran, Dhami, Schramowski, and Kersting]{shindo2024deisam}
Hikaru Shindo, Manuel Brack, Gopika Sudhakaran, Devendra~Singh Dhami, Patrick Schramowski, and Kristian Kersting.
\newblock Deisam: Segment anything with deictic prompting.
\newblock \emph{arXiv preprint arXiv:2402.14123}, 2024.

\bibitem[Sun et~al.(2019)Sun, Myers, Vondrick, Murphy, and Schmid]{sun2019videobert}
Chen Sun, Austin Myers, Carl Vondrick, Kevin Murphy, and Cordelia Schmid.
\newblock Videobert: A joint model for video and language representation learning.
\newblock In \emph{Proceedings of the IEEE/CVF international conference on computer vision}, pp.\  7464--7473, 2019.

\bibitem[Sun et~al.(2023)Sun, Bao, Liu, Mei, and Black]{sun2023trace}
Yu~Sun, Qian Bao, Wu~Liu, Tao Mei, and Michael~J. Black.
\newblock Trace: 5d temporal regression of avatars with dynamic cameras in 3d environments.
\newblock \emph{arXiv preprint arXiv:2306.02850}, 2023.

\bibitem[Thomee et~al.(2016)Thomee, Shamma, Friedland, Elizalde, Ni, Poland, Borth, and Li]{thomee2016yfcc100m}
Bart Thomee, David~A Shamma, Gerald Friedland, Benjamin Elizalde, Karl Ni, Douglas Poland, Damian Borth, and Li-Jia Li.
\newblock Yfcc100m: The new data in multimedia research.
\newblock \emph{Communications of the ACM}, 59\penalty0 (2):\penalty0 64--73, 2016.

\bibitem[Xie et~al.(2018)Xie, Sun, Huang, Tu, and Murphy]{xie2018s3d}
Saining Xie, Chen Sun, Jonathan Huang, Zhuowen Tu, and Kevin Murphy.
\newblock Rethinking spatiotemporal feature learning: Speed-accuracy trade-offs in video classification.
\newblock In \emph{Proceedings of the European conference on computer vision}, pp.\  305--321, 2018.

\bibitem[Xu et~al.(2021)Xu, Ghosh, Huang, Okhonko, Aghajanyan, Metze, Zettlemoyer, and Feichtenhofer]{xu2021videoclip}
Hu~Xu, Gargi Ghosh, Po-Yao Huang, Dmytro Okhonko, Armen Aghajanyan, Florian Metze, Luke Zettlemoyer, and Christoph Feichtenhofer.
\newblock Videoclip: Contrastive pre-training for zero-shot video-text understanding.
\newblock \emph{arXiv preprint arXiv:2109.14084}, 2021.

\bibitem[Xu et~al.(2022{\natexlab{a}})Xu, Qu, Kuen, Gu, and Liu]{xu2022meta}
Li~Xu, Haoxuan Qu, Jason Kuen, Jiuxiang Gu, and Jun Liu.
\newblock Meta spatio-temporal debiasing for video scene graph generation.
\newblock \emph{arXiv preprint arXiv:2207.11441}, 2022{\natexlab{a}}.

\bibitem[Xu et~al.(2022{\natexlab{b}})Xu, Rawat, Wong, Kankanhalli, and Shah]{xu2022dont}
Ziwei Xu, Yogesh~S Rawat, Yongkang Wong, Mohan Kankanhalli, and Mubarak Shah.
\newblock Don't pour cereal into coffee: Differentiable temporal logic for temporal action segmentation.
\newblock In Alice~H. Oh, Alekh Agarwal, Danielle Belgrave, and Kyunghyun Cho (eds.), \emph{Advances in Neural Information Processing Systems}, 2022{\natexlab{b}}.

\bibitem[Yang et~al.(2018)Yang, Lu, Lee, Batra, and Parikh]{yang2018graph}
Jianwei Yang, Jiasen Lu, Stefan Lee, Dhruv Batra, and Devi Parikh.
\newblock Graph r-cnn for scene graph generation.
\newblock In \emph{Proceedings of the European conference on computer vision}, pp.\  670--685, 2018.

\bibitem[Yang et~al.(2023{\natexlab{a}})Yang, Peng, Li, Guo, Chen, Li, Ma, Zhou, Zhang, Loy, and Liu]{yang2023panoptic}
Jingkang Yang, Wenxuan Peng, Xiangtai Li, Zujin Guo, Liangyu Chen, Bo~Li, Zheng Ma, Kaiyang Zhou, Wayne Zhang, Chen~Change Loy, and Ziwei Liu.
\newblock Panoptic video scene graph generation.
\newblock \emph{arXiv preprint arXiv:2311.17058}, 2023{\natexlab{a}}.

\bibitem[Yang et~al.(2023{\natexlab{b}})Yang, Ma, Huang, Chen, Lin, Han, and Chang]{yang2023tempclr}
Yuncong Yang, Jiawei Ma, Shiyuan Huang, Long Chen, Xudong Lin, Guangxing Han, and Shih-Fu Chang.
\newblock Tempclr: Temporal alignment representation with contrastive learning.
\newblock \emph{arXiv preprint arXiv:2212.13738}, 2023{\natexlab{b}}.

\bibitem[Yao et~al.(2021)Yao, Zhang, Han, Li, Weber, Liu, Wermter, and Sun]{Yao2021ICCV}
Yuan Yao, Ao~Zhang, Xu~Han, Mengdi Li, Cornelius Weber, Zhiyuan Liu, Stefan Wermter, and Maosong Sun.
\newblock Visual distant supervision for scene graph generation.
\newblock In \emph{Proceedings of the IEEE/CVF International Conference on Computer Vision}, pp.\  15816--15826, October 2021.

\bibitem[Zhai et~al.(2023)Zhai, Mustafa, Kolesnikov, and Beyer]{zhai2023sigmoid}
Xiaohua Zhai, Basil Mustafa, Alexander Kolesnikov, and Lucas Beyer.
\newblock Sigmoid loss for language image pre-training.
\newblock \emph{arXiv preprint arXiv:2303.15343}, 2023.

\bibitem[Zhang et~al.(2023)Zhang, Pan, Yao, Huang, Mei, and Chen]{zhang2023learning}
Yong Zhang, Yingwei Pan, Ting Yao, Rui Huang, Tao Mei, and Chang-Wen Chen.
\newblock Learning to generate language-supervised and open-vocabulary scene graph using pre-trained visual-semantic space.
\newblock In \emph{Proceedings of the IEEE/CVF Conference on Computer Vision and Pattern Recognition}, pp.\  2915--2924, 2023.

\bibitem[Zhang et~al.(2024)Zhang, Wu, Li, Li, Ma, Liu, and Li]{zhang2024videoinstructiontuningsynthetic}
Yuanhan Zhang, Jinming Wu, Wei Li, Bo~Li, Zejun Ma, Ziwei Liu, and Chunyuan Li.
\newblock Video instruction tuning with synthetic data, 2024.
\newblock URL \url{https://arxiv.org/abs/2410.02713}.

\bibitem[Zhu et~al.(2022)Zhu, Zhang, Jiang, Dang, Hou, Shen, Feng, Zhao, Miao, Shah, et~al.]{zhu2022scene}
Guangming Zhu, Liang Zhang, Youliang Jiang, Yixuan Dang, Haoran Hou, Peiyi Shen, Mingtao Feng, Xia Zhao, Qiguang Miao, Syed Afaq~Ali Shah, et~al.
\newblock Scene graph generation: A comprehensive survey.
\newblock \emph{arXiv preprint arXiv:2201.00443}, 2022.

\end{thebibliography}

\newpage
\appendix

\pagebreak
\section{Natural Language to STSL Specification}
\label{app:nl2spec}

We designed a generic prompt template to convert a caption into its structured representation. The prompt template consists of the following components:
\begin{enumerate}[topsep=0pt,itemsep=-1ex,partopsep=1ex,parsep=1ex,leftmargin=0.5cm]
\item Set the LLM to be ``a super user in logic programming''.
\item Define video location and duration as fractions of the whole video length.
\item Include desired scene graph keywords, such as object names and relations.
\item Provide n-examples of captions and their corresponding JSON structured representations.
\end{enumerate}
We include the full prompts for OpenPVSG dataset and 20BN-Something-Something dataset here.

\subsection{Character Setup}
We setup the agent character for all the datasets with this prompt.

\begin{lstlisting}[frame=single]
You are a super user in logic programming. 
\end{lstlisting}

\subsection{Constant definition}

We setup the examples for temporal specification for all datasets.
\begin{lstlisting}[frame=single]
Examples of precise video locations: 1/4, 1/2, 2/3, 1.
Examples of event durations: 1/4, 2/3, 1.
\end{lstlisting}

We give the concept of ``unary predicate'' and ``binary predicate'' for all datasets.
\begin{lstlisting}[frame=single]
A static predicate represents the property of the target object. 
For example, is_tearable(A) means the A can be teared. 
A unary predicate takes in one argument. 
For example, close(A) means A is close to the camera.
A binary predicate takes in two arguments. 
For example, above(A, B) means A is above B.
\end{lstlisting}

\subsection{Scene Graph Vocabularies}
To bias the GPT-4 model to generate more related words align with the desired scene graph vocabulary, we added the OpenPVSG vocabularies into the prompt as well. 

We show the scene graph vocabulary part of OpenPVSG prompt below.
\begin{lstlisting}[frame=single]
The entites in the video can be: 
    adult, baby, bag, ..., tree, wall, water
The relations in the video can be: 
    beside, biting,  blowing, brushing, ..., watering, wearing
\end{lstlisting}

We show the scene graph vocabulary part of 20BN-Something-Something prompt below.
\begin{lstlisting}[frame=single]
Static predicates: 
    is_bendable, is_fluid, is_holdable, is_rigid, is_tearable, neq
Unary predicates: 
    broken, close, closed, deformed, empty, far, ..., twisted, upright
Binary predicates: 
    above, attached, behind, fits, in, ..., touching, under, visible
Constants: 
    hand
\end{lstlisting}

\subsection{Few-shot Example}
In both cases, we provide the prompt with 3 examples converting the caption into a structured representation with json format.
The target json file includes a detailed version of caption, the programmatic version in logical predicates, video location of the event and the duration of the event. 

We show one example mapping the caption into json format below for OpenPVSG dataset below.

\begin{lstlisting}[frame=single,basicstyle=\scriptsize\tt]
Caption: "The young boy receives another gift and sits on the floor."
Action json:
{
    "caption": "The young boy receives another gift and sits on the floor.",
    "sequential descriptions": [
        "boy A receives gift B",
        "boy A sits on the floor C", 
        ],
    "time stamps": {
        "1": {
            "decription": [
                "boy A receives gift B",
            ],
            "programmatic": [
                "holding(B, D)",
                "name(A, boy)",
                "name(B, gift)",
            ],
            "duration": "1/2",
            "video location": "0"
        },
        "2": {
            "decription": [
                "boy A sits on the floor C",
            ],
            "programmatic": [
                "sitting on(A, C)",
                "name(A, boy)",
                "name(C, floor)",
            ],
            "duration": "1/3",
            "video location": "1"
        }
    }
}

\end{lstlisting}

We show one example mapping the caption into json format below for 20BN dataset below.

\begin{lstlisting}[frame=single,basicstyle=\scriptsize\tt]
Action: dig
Action predicate: dig(A, B)
Action description: Digging [A] out of [B].
Action json:
{
    "action": "dig",
    "action_pred": "dig(A, B)",
    "explanation": "Digging [A] out of [B]",
    "static properties": [
        "A is holdable", 
        "B is not rigid", 
        "entity A and B are not equvalent"
    ],
    "programmatic version": [
        "is_holdable(A)", 
        "not(is_rigid(B))",
        "neq(A, B)"
    ],
    "time stamps": {
        "1": {
            "decription": [
                "Not A and B are far away from the camera",
                "A is in B",
                "A and B are not touched by hand", 
                "A is not visible, but B is visible,
                and there is a hand that is visible"
            ],
            "programmatic": [
                "not(far(A)), not(far(B))",
                "in(A, B)",
                "not(touching(A, hand)), not(touching(B, hand))",
                "not(visible(A)), visible(B), visible(hand)"
            ],
            "duration": "1/5",
            "video location": "0"
        },
        "2": {
            "decription": [
                "A become visible", 
                "A is in a hand", 
                "A is not touching B"
            ],

            "programmatic": [
                "visible(A)", 
                "in(A, hand)", 
                "not(touching(A, B))"
            ],
            "duration": "1/5",
            "video location": "1"
        }
    },
}
\end{lstlisting}

We have implemented corresponding post-processors to parse the json structures into STSL programs.
\section{Experimental Details}
\label{app:experimental_details}
\subsection{Hardware}
All experiments are conducted on two machines:
\begin{enumerate}
    \item  two 20-core Intel Xeon CPUs, four GeForce RTX 2080 Ti GPUs, and 768 GB RAM,
    \item 3.00GHz Intel Xeon machine with 48 CPUs, 8 A100s, and 1.5T RAM.
\end{enumerate}

\subsection{OpenPVSG}
\label{app:openpvsg_experiment}

\textbf{Dataset.}
The OpenPVSG \cite{yang2023panoptic} dataset comprises $400$ videos sourced from Ego4D\cite{grauman2021ego4d}, VidOr\cite{shang2019annotating, thomee2016yfcc100m}, and EpicKitchen\cite{Damen2022RESCALING, Damen2018EPICKITCHENS}. 
This dataset offers fine-grained ground truth annotations of STSGs for $150K$ frames. 
These scene graphs encompass $126$ object classes and $57$ relation classes, including pixel-level object masks.
Each video is accompanied by multiple captions with their corresponding video snippets. 
The training dataset contains $1,538$ caption-video pairs, while the testing dataset includes $295$ caption-video pairs. 
We train and test on these caption-video snippet pairs.

\textbf{Experimental Setup.}
As illustrated in Figure \ref{fig:model-architecture}, our objective is to train an STSG extraction model capable of taking a video clip along with object bounding boxes as input and predicting the properties, attributes of objects, and relationships between objects.
As the ground truth programmatic specification is not provided in the OpenPVSG dataset, we only adopt the \ours-\textbf{C} architecture. 
The 3-shot GPT-4 model processes the captions and converts them into structured representations of events with logical predicates $\psi_i$.
By postprocessing, we extract all the unary and binary keywords from the GPT-4 generated $\psi$, and use them as input to the STSG extraction model.
We finetune the SigLIP model to generate the probabilistic spatio-temporal scene graph.
This model predicts both the similarity scores between the cropped objects and the unary predicate keywords, and the similarity scores between object pairs and the binary predicate keywords, yielding a probabilistic STSG.
Due to SigLIP's characteristic as an embedding-based model, it supports open-world vocabulary and can handle the fuzziness that may be present in the GPT-4 generated structured representations.
Due to the hardware limitation, we set the batch size to 1, and our contrastive loss is obtained by randomly sampling another negative specification other than the current specification.

\textbf{Evaluation Metrics.}
We use recall@k for evaluating the unary and binary STSG generating performance. 
The recall metric checks given the ground truth bounding boxes and object pairs, whether the predicted unary predicate, in this case, object names, ranks within the top-k  among the ground truth unary predicates.
Then we compare the accuracy against the ground truth binary labels. 

To obtain the evaluation results for the OpenPVSG baselines, the object identification results need to be isolated from the entire prediction pipeline. The IPS-based OpenPVSG models first predict a class label for each pixel, then aggregate similar predictions into a single object using Unitrack. In contrast, the VPS-based models directly predict object tubes with corresponding labels. To separate object identification from the segmentation task, we identify predicted objects that sufficiently overlap with ground truth objects (with an IOU greater than 0.5). We then calculate the average label for each predicted object across the entire video. The label probability is determined by the proportion of frames in which the object is assigned a particular label, relative to the total frames in which the object appears.

\textbf{Learning Setup}
For STSG generator,  we finetune the ``google/siglip-base-patch16-224'', ``openai/clip-vit-large-patch14'', and the pretrained violet model checkpoint with learning rate $0.000001$ for $10$ epoches.
To ensure the inference efficiency, we adopt sampling technique, uniformly sample the whole video frames, and ensure the final generated video has at most $12$ frames. 
Further, we preserve only $10$ largest objects on the image, and for each frame, we sample $100$ relation pairs at training time, and $300$ relation pairs at test time. 
We set the learning batch size to $1$.

\subsection{LLaVA-Video Zero-Shot Transferability}
We trained CLIP model with LASER method on 10K LLaVA-Video dataset \cite{zhang2024videoinstructiontuningsynthetic}, and test on OpenPVSG evaluation dataset, showing generalizability, robustness, and scalability.

\subsubsection{Dataset Preprocessing with LLM + Compiler Pipeline }

\textbf{OpenPVSG Caption to STSL Program}
The OpenPVSG captions average 12.69 words, containing an average of 1.69 events in the GPT-extracted representations.
The failure rate for converting captions into valid executable STSL formulas is 0.78\%.
A total of 453 unique unary keywords and 679 unique binary keywords were identified, covering:
92.86\% ground-truth unary and 94.74\% ground-truth binary keywords in the OpenPVSG dataset.
55.56\% ground-truth unary and 61.54\% ground-truth binary keywords in the Action Genome dataset.
20\% ground-truth unary and 12.88\% ground-truth binary keywords in the VidVRD dataset.
Notably, these closed-world datasets include keywords like ``unsure,'' ``other relation,'' and ``not contacting,'' which rarely occur in natural scenes.

\textbf{LLaVA-Video 10K Dataset Caption to STSL Program}
To demonstrate the capability of our LLM + compiler pipeline in handling long and complex caption descriptions, we evaluated it on LLaVA-Video, a dataset with detailed captions released on October 4, 2024. 
We study the quality of the extracted captions on a randomly sampled subset of 10,000 video clips (each under 30 seconds). 
The LLaVA-Video 10K captions average 233.05 words, and the pipeline extracted 4.03 events per video on average.
The failure rate for converting captions into valid executable STSL formulas was 0\%.
A total of 18,458 unique unary keywords and 4,492 unique binary keywords were identified, covering:
91.27\% ground-truth unary and 89.47\% ground-truth binary keywords in the OpenPVSG dataset.
80.56\% ground-truth unary and 69.23\% ground-truth binary keywords in the Action Genome dataset.
82.86\% ground-truth unary and 31.82\% ground-truth binary keywords in the VidVRD dataset.
Notably, 91.99\% of LLaVA-Video samples originate from YouTube, with the remaining samples sourced from ActivityNet, Charades, Ego4D, NextQA, and YouCook2. This highlights a significant domain shift between the LLaVA-Video dataset and all evaluation datasets.

\subsubsection{Keyword Analysis}
We present the top 10 most frequent unary and binary keywords extracted from captions in both datasets to further illustrate the diversity and robustness of our method.

\begin{table}[tbh!]
    \centering
    \renewcommand{\arraystretch}{1.2} % Adjust row height
    \setlength{\tabcolsep}{6pt} % Adjust column spacing
    \begin{tabular}{l c l c l c l c}
        \toprule
        \multicolumn{2}{c}{LLaVA-Video Unary} & \multicolumn{2}{c}{LLaVA-Video Binary} & \multicolumn{2}{c}{OpenPVSG Unary} & \multicolumn{2}{c}{OpenPVSG Binary} \\
        \cmidrule(r){1-2} \cmidrule(r){3-4} \cmidrule(r){5-6} \cmidrule(r){7-8}
        Category & Count & Category & Count & Category & Count & Category & Count \\
        \midrule
        women    & 1020  & hold       & 2531  & adult    & 823  & on         & 264  \\
        hand     & 886   & wear       & 2275  & child    & 569  & holding    & 211  \\
        man      & 863   & on         & 1103  & man      & 322  & picking    & 163  \\
        text     & 499   & in         & 811   & ball     & 254  & placing    & 148  \\
        child    & 421   & with       & 725   & I        & 206  & using      & 135  \\
        camera   & 419   & adjust     & 386   & dog      & 187  & toward     & 113  \\
        character& 414   & color      & 362   & toy      & 127  & in         & 90   \\
        hands    & 256   & stand      & 346   & woman    & 111  & throwing   & 83   \\
        room     & 243   & near       & 286   & Baby     & 106  & playing    & 66   \\
        object   & 239   & sit        & 275   & camera   & 104  & sitting on & 65   \\
        \bottomrule
    \end{tabular}
    \caption{Comparison of LLaVA-Video and OpenPVSG categories and their counts.}
    \label{tab:comparison}
\end{table}

\subsubsection{Cost Analysis}
Cost of Generating the STSL from Intermediate Representations
Generating all intermediate representations with GPT for the $10K$ dataset costs approximately \$$50$ and takes around $15$ minutes using parallelization techniques.
The compilation process from the $10K$ GPT-generated structured representations to executable STSL programs requires about 1 minute.

\subsection{Zero-shot Tranferability}
Training with Noisy Object Trajectories and Long Captions on LLaVA-Video
As the LLaVA-Video dataset does not provide ground truth object mask-level trajectories. To address this, we preprocess the videos using SAM2.1 to extract object trajectories and train our model with noisy object trajectories and weak supervision labels. A uniform set of hyperparameters for the SAM2.1 mask generator was determined through grid search, and mask quality was manually verified. On average, we extract $19.87$ object trajectories for each video, and on average $10.61$ objects occur on a single frame. Compared to the OpenPVSG dataset, there are $20.47$ object trajectories for each video, and on average $10.36$ objects occur on a single frame. For the Action Genome and VidVRD datasets, only coarse-grained bounding boxes are available. During evaluation, these bounding boxes are converted into masks by setting all parts within the bounding boxes to True and areas outside to False.

The chart below illustrates the learning performance of our method, leveraging CLIP as the backbone. The results demonstrate that our approach is robust in handling complex caption descriptions, noisy object trajectories and out-of-domain transfer scenarios.

\begin{table}[h]
    \centering
    \renewcommand{\arraystretch}{1.2} % Adjust row height
    \setlength{\tabcolsep}{5pt} % Adjust column spacing
    \begin{tabular}{l l c c c c c c}
        \toprule
        \multicolumn{2}{c}{Eval Dataset \& LASER FT Strategy} & \multicolumn{3}{c}{Unary R@K} & \multicolumn{3}{c}{Binary R@K} \\
        \cmidrule(r){3-5} \cmidrule(r){6-8}
        Dataset & Strategy & R@1 & R@5 & R@10 & R@1 & R@5 & R@10 \\
        \midrule
        \multirow{3}{*}{OpenPVSG} 
        & Base & 0.1633 & 0.3381 & 0.4404 & 0.0197 & 0.0673 & 0.0988 \\
        & LLaVA-FT & 0.2368 & 0.5000 & 0.5789 & 0.1191 & 0.3534 & 0.5346 \\
        & OpenPVSG-FT & 0.2778 & 0.5231 & 0.6402 & 0.1482 & 0.4214 & 0.5398 \\
        \midrule
        \multirow{3}{*}{Action Gnome} 
        & Base & 0.1487 & 0.4166 & 0.5911 & 0.0509 & 0.2464 & 0.5478 \\
        & LLaVA-FT & 0.1768 & 0.4795 & 0.6471 & 0.1255 & 0.5250 & 0.6772 \\
        & OpenPVSG-FT  & 0.1422 & 0.4367 & 0.6277 & 0.2407 & 0.3085 & 0.6927 \\
        \midrule
        \multirow{3}{*}{VidVRD} 
        & Base & 0.6267 & 0.8408 & 0.9332 & 0.0499 & 0.1696 & 0.2445 \\
        & LLaVA-FT & 0.6444 & 0.8899 & 0.9585 & 0.0678 & 0.1979 & 0.2966 \\
        & OpenPVSG-FT & 0.5925 & 0.8322 & 0.9161 & 0.0211 & 0.1179 & 0.1715 \\
        \bottomrule
    \end{tabular}
    \caption{Performance comparison of LASER finetuning strategies across different evaluation datasets. Note that the LLaVA-FT is out of distribution and OpenPVSG-FT is in domain. }
    \label{tab:laser_ft_clip}
\end{table}

\subsubsection{Using STSL as STSG}
We conducted an experiment on the OpenPVSG eval split, comparing the performance of a model directly using the generated STSL data (referred to as STSL-as-STSG) against our LASER fine-tuned CLIP-based STSG model. Here is the detailed experimental setup, partially addressing the aforementioned challenges:

Vocabulary Alignment: To address challenge (a), we use the spaCy package to project caption-extracted STSL keywords into the dataset-specific vocabulary. When a direct match was not found, we took the top-10 predictions based on similarity scores and assigned the similarity score as the predicted likelihood.
Assuming Perfect Grounding: To address challenge (b) and (c), we assumed a hypothetical, ideal algorithm that grounds entities and their relations into specific frames of the video perfectly, bypassing errors related to spatial-temporal grounding.
Evaluation Metric An object is counted as a match for unary if its name appears in the top-k most likely predictions of the projected STSL within its predicted event duration. Similarly, a pair of objects is counted as a match for binary if their binary relation is among the top-k most likely predictions of the projected STSL within the predicted event duration.
Even under these over-optimistic assumptions, our experiment revealed that a model directly using STSL data without fine-tuning performs significantly worse compared to the fine-tuned CLIP-based STSG model. Through our investigation, we found that the performance degradation is primarily due to (d) mismatched granularity, which cannot be easily addressed without fine-tuning. As a side note, our fine-tuned STSG model produces results much faster than STSL-as-STSG during inference time. On average, calling the full STSL-as-STSG pipeline takes about $34.5$ seconds per video for extracting the STSL from the video, while directly calling the fine-tuned CLIP model takes about $2.6$ seconds per video on the OpenPVSG dataset.

\begin{table}[h]
    \centering
    \renewcommand{\arraystretch}{1.2} % Adjust row height
    \setlength{\tabcolsep}{5pt} % Adjust column spacing
    \begin{tabular}{l l c c c c c c}
        \toprule
        \multicolumn{2}{c}{Eval Dataset \& LASER FT Strategy} & \multicolumn{3}{c}{Unary R@K} & \multicolumn{3}{c}{Binary R@K} \\
        \cmidrule(r){3-5} \cmidrule(r){6-8}
        Dataset & Strategy & R@1 & R@5 & R@10 & R@1 & R@5 & R@10 \\
        \midrule
        \multirow{4}{*}{OpenPVSG} 
        & Base & 0.1633 & 0.3381 & 0.4404 & 0.0197 & 0.0673 & 0.0988 \\
        & LLaVA-FT & 0.2368 & 0.5000 & 0.5789 & 0.1191 & 0.3534 & 0.5346 \\
        & OpenPVSG-FT & 0.2778 & 0.5231 & 0.6402 & 0.1482 & 0.4214 & 0.5398 \\
        & (NEW) STSL-as-STSG & 0.0514 & 0.1132 & 0.1288 & 0.0362 & 0.0583 & 0.0716 \\
        \bottomrule
    \end{tabular}
    \caption{Performance comparison of LASER FT CLIP strategies on the OpenPVSG dataset.}
    \label{tab:laser_ft_clip_openpvsg}
\end{table}

\subsubsection{State-of-the-art Fully Supervised Baselines}

While STTran and TRACE demonstrate strong performance in video scene graph generation, they are closed-domain models with fixed classification layers, limiting their ability to transfer knowledge across datasets. Additionally, their architectures are not designed to handle noisy labels directly extracted from captions, making them less robust in such scenarios.

It is important to note that the primary contributions of STTran and TRACE lie in their carefully designed model architectures for capturing visual features. In contrast, our work focuses on an orthogonal problem: developing a model-agnostic weak supervision training pipeline. Our approach is not in competition with STTran or TRACE but rather complements them. In fact, STTran and TRACE can be adopted as backbone models within our LASER framework, demonstrating that our method enhances their capabilities rather than being mutually exclusive.

Nevertheless, we report unary and binary predicate recall for STTran using the author-provided checkpoint. Note that STTran is trained in-domain on the Action Genome dataset in a fully supervised manner, while our models are trained on out-of-domain datasets using a weakly supervised approach.

\begin{table}[h]
    \centering
    \renewcommand{\arraystretch}{1.2} % Adjust row height
    \setlength{\tabcolsep}{5pt} % Adjust column spacing
    \begin{tabular}{l l c c c c c c}
        \toprule
        \multicolumn{2}{c}{Eval Dataset \& Strategy} & \multicolumn{3}{c}{Unary R@K} & \multicolumn{3}{c}{Binary R@K} \\
        \cmidrule(r){3-5} \cmidrule(r){6-8}
        Dataset & Strategy & R@1 & R@5 & R@10 & R@1 & R@5 & R@10 \\
        \midrule
        \multirow{4}{*}{Action Gnome} 
        & Base CLIP & 0.1487 & 0.4166 & 0.5911 & 0.0509 & 0.2464 & 0.5478 \\
        & \shortstack{LLaVA-FT} & 0.1768 & 0.4795 & 0.6471 & 0.1255 & 0.5250 & 0.6772 \\
        & \shortstack{OpenPVSG-FT} & 0.1422 & 0.4367 & 0.6277 & 0.2407 & 0.3085 & 0.6927 \\
        & \shortstack{STTran } & 0.0671 & 0.2879 & 0.4632 & 0.4097 & 0.7815 & 0.8886 \\
        \bottomrule
    \end{tabular}
    \caption{Performance comparison of different strategies on the Action Gnome dataset. We note that LLaVA-FT and OpenPVSG-FT is out of distribution, and weakly supervised, while STTran is in domain and fully supervised.}
    \label{tab:action_gnome}
\end{table}

\subsection{MUGEN}
\label{app:mugen_experiment}
\begin{figure}[tbh!]
    \begin{subfigure}[b]{.45\linewidth}
    \includegraphics[width=\linewidth]{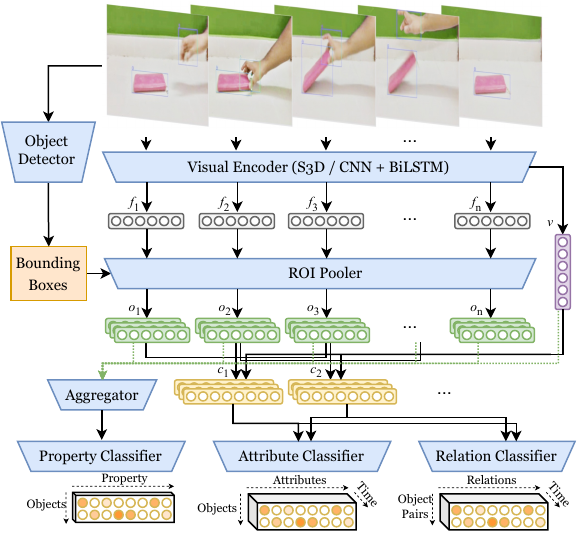}
    \caption{
    20BN-Something-Something classifier model architecture used in \ours-P.
    }
    \label{fig:20bn_architecture}
    \end{subfigure}
\hfill
    \begin{subfigure}[b]{.45\linewidth}
    \includegraphics[width=\linewidth]{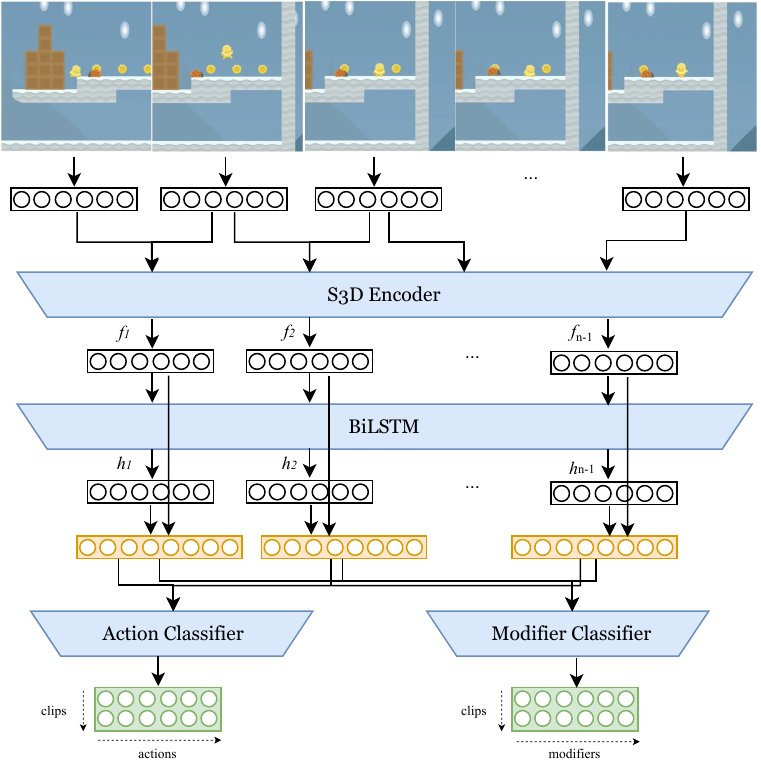}
    \caption{
    MUGEN model classifier architecture used in \ours-P.
    }
    \label{fig:mugen_architecture}
    \end{subfigure}
\end{figure}

\begin{figure}[htb]
    \centering
    \includegraphics[width=0.5\linewidth]{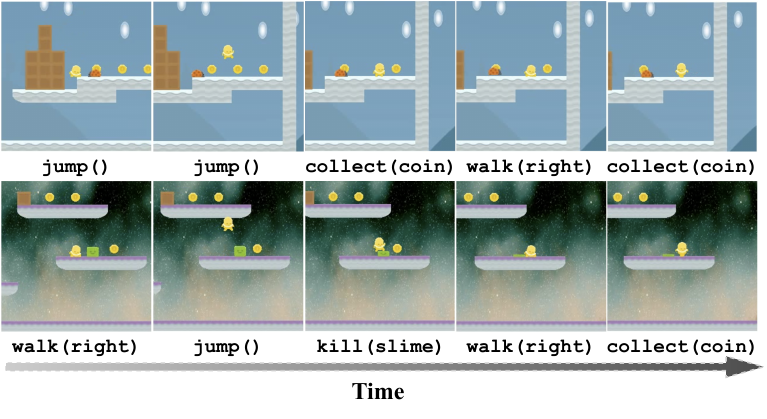}
    \caption{
    Examples of the predicted MUGEN actions predicted. 
    We elide showing \texttt{M} in each predicate for brievity.
    }
    \label{fig:mugen_qualitative}
\end{figure}

\begin{table}[tbh!]
\centering
    \begin{tabular}{ c||c|c  } 
    \toprule
    \hline
    \textbf{Task} & \textbf{SDSC} & \oursbf-\textbf{P} \\
    \hline
    Video-to-Text Retrieval & 87.00\% & \textbf{93.80\%} \\
    Text-to-Video Retrieval & 86.80\% & \textbf{90.00\%} \\
    \bottomrule
    \end{tabular}
    \caption{
    Comparison to baselines on downstream retrieval tasks with the MUGEN dataset.
    }
    \label{tab:downstream}
\end{table}

\textbf{Model Architecture}
We present the model architecture overview in \figref{fig:mugen_architecture}.
The video is first segmented into clips through a sliding window of length 2. 
Then, each clip is encoded through an S3D model and yields clip-wise embedding. 
The embedding is further passed into a BiLSTM model to obtain the context for each clip.
An action classifier takes in the concatenation of the clip-based embedding and its context embedding, and classifies each clip into $6$ actions: 
\texttt{walk}, \texttt{jump}, \texttt{kill}, \texttt{collect}, \texttt{die}, and \texttt{climb}.
Further, a modifier classifier predicts the $4$ possible direction of the actions: \texttt{left}, \texttt{right}, \texttt{up}, \texttt{down}.
The matching process of action and its modifier is also performed in the reasoner. 
For example, a combination of \texttt{collect}, \texttt{left} will be invalidated during reasoning. 

\textbf{Learning Setup}
We train the model on $5000$ training datapoints, and $12,851$ test datapoints. 
The contrastive loss is obtained over a batch size 3; the violation loss is constructed with regard to the axioms in GPA \cite{migimatsu2022grounding}.
We train the model with a learning rate of $0.0001$, violation weight $0.01$, number of epochs $100$, and batch size $3$.
The Scallop reasoning engine setup is using difftopkproofs provenance with $k$ set to $5$.
We construct the temporal specification from the given text description using a simple heuristic-based semantic parser. 
We first extract an ordered action list $[a_1, a_2, \dots, a_n]$ from the text with heuristics, then construct the temporal specification as $a_1  \mathbf{U} a_2  \dots \mathbf{U} a_n$.
This specification means the actions are performed one by one with no intermediate gaps.
The generated specifications are used as ground truth programmatic STSL labels.
As the MUGEN video data is very different from the natural videos, we train a classification model from scratch.
Our neural perception model consists of an S3D video encoding model pretrained on Kinetics 400 \cite{kay2017kinetics}, and MLP layers to classify the actions.
The video is first passed through the S3D \cite{xie2018s3d} model for frame-based embeddings. 
Then all the embeddings that occur in a video clip are concatenated as the input to a $2$-layer MLP classifier.
We set the number of frames per clip to $2$, and the batch size to $3$ due to the hardware limitation.
The classifier thus produces clip-wise video representations.
We employ contrastive loss and semantic loss for training.
More details are shown in \appref{app:mugen_experiment}.

\textbf{Quantitative Study.}
We evaluate our action prediction performance on MUGEN and compare it to caption supervised VT-TWINS \cite{ko2022video}, TempCLR \cite{yang2023tempclr}, and a directly supervised baseline (Supervised). 
We provide all baselines with extra annotations on the start- and end-frame of each clip-based textual description.
As the ground truth actions for the video are different from the text description, we use heuristics to obtain the start and end frames.
For a fair comparison, we fine-tune an S3D backbone pre-trained using Kinetics 400 (the same as \ours) along with the baseline models.
This label is used for training both VT-TWINS and fully supervised models.
As shown in Table \ref{tab:performance}, \ours~has better action prediction accuracy despite receiving less supervision than the baselines.
Moreover, our weakly supervised model even achieves better accuracy than the Supervised method on \#Data $= 5000$.
We further evaluate our approach on a downstream video-specification retrieval task. 
Given 3 videos and 3 specifications, we want to match the correctly aligned pairs, retrieve specification given video and vice versa.
We denote the two tasks as spec-retrieval and video-retrieval respectively.
Specifically, we evaluate the accuracy with which we infer whether the correct specification has the highest alignment score, and vice versa.
\ours{} outperforms an embedding-based baseline SDSC \cite{hayes2022mugen} on both tasks (Table~\ref{tab:downstream}).
Our approach can even identify actions that persist for a very short period of time, such as \texttt{kill} an enemy.

\subsection{20BN-Something-Something.}
\begin{figure*}[tbh!]
    \centering
    \includegraphics[width=\linewidth]{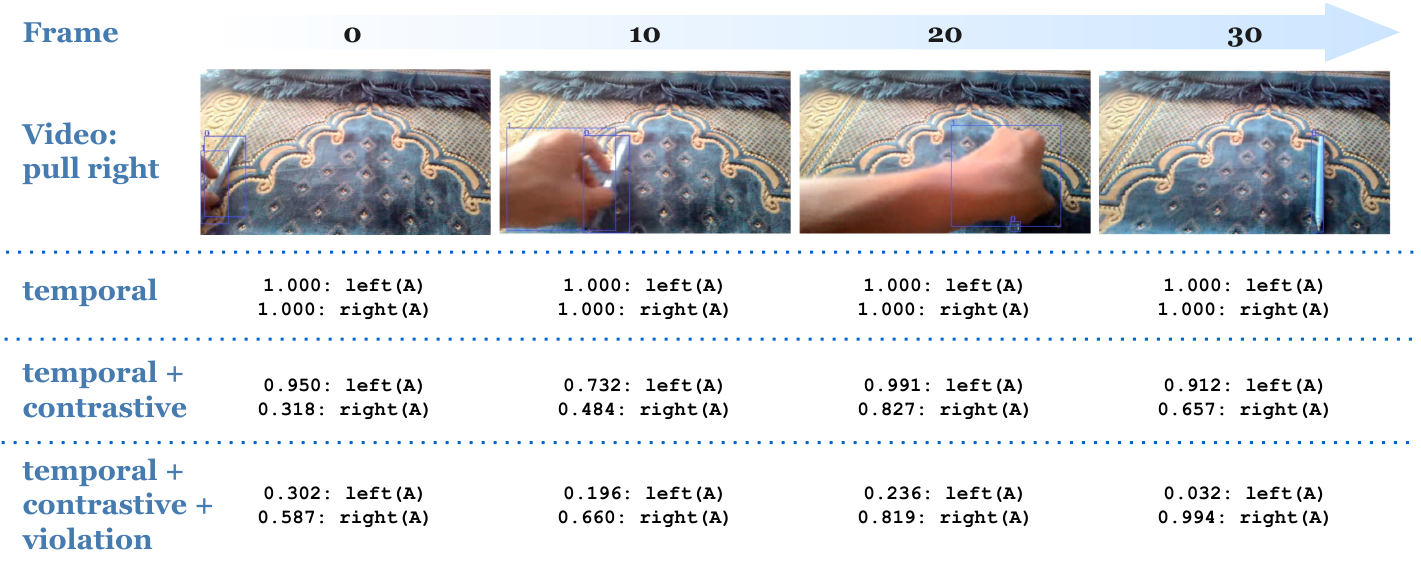}
    \caption{
    Comparing model performance trained with different loss functions.
    }
    \label{fig:20bn-ablation}
\end{figure*}

\label{app:20bn_details}
\textbf{Dataset Details}
We trained on $10,000$ datapoints and tested on $14,816$ datapoints.
There are 172 actions with 37 underlying predicates in this dataset.
Specifically, there are 6 static predicates, 21 unary predicates, and 10 binary predicates.
The static predicates represent object attributes that persist over time;
the unary predicates reflect a single object's state;
the binary predicates reflect the relationship between two objects.

\textbf{Model Architecture}
We use a 4-layer convolutional model to extract the features for each frame and a ROIpooler to obtain the embedding for each object on the frame.
To obtain the static predicates, we pass the frame-wise object embeddings through an LSTM encoder to obtain the video-wise object embedding, and a 2-layer MLP classifier yields the output distribution.
For the unary predicates, a 2-layer MLP classifier takes in the concatenation of the frame embedding, object embedding, and the object bounding box, and generates the output distribution.
For the binary predicates, a 3-layer MLP classifier takes in the concatenation of the frame embedding, two object embeddings, and two object bounding boxes, and generates the output distribution.
The overview of the architecture is shown in \figref{fig:model-architecture}.

We use a convolutional model for the frame features, an ROI Pooler for object features, an LSTM model for video-wise object embeddings, and MLP layers for predicate output distributions.
We adopt temporal supervision loss with the prior that the precondition and postcondition should be far away from each other;  contrastive learning loss with batch size 3; and semantic loss with a weight of 0.05, where the integrity constraints, such as ``a rigid object is not fluid'', are obtained from the original PDDL file. 
More training details are included in the appendix.

Instead of using just querying for the alignment score $Pr(\mathbf{r} \models \psi)$, we instead query for the conditional alignment score $Pr(\mathbf{r} \models \psi | d)$ which is conditioned on the distance $d$ between $\psi_{\texttt{pre}}$ and $\psi_{\text{post}}$ are satisfied. 
% The full loss function is shown in \eqnref{eqn:20bn-loss-fn}.
In the experiment, we set the threshold $d_{\text{min}}$ to be $0.9d_{\text{max}}$.

\textbf{Learning Setup.}
The contrastive loss is obtained over a batch size 3; the violation loss is constructed with regard to the axioms in GPA \cite{migimatsu2022grounding}.
We train the model with a learning rate of $0.0001$, violation weight $0.05$, number of epochs $50$, and batch size $3$.
The Scallop reasoning engine setup is using difftopkproofs provenance with k=3.

\textbf{Ablation Studies.}
We study how different loss impact the qualitative evaluation result, as shown in \figref{fig:20bn-ablation}.
In this task, the spatiotemporal specifications are manually crafted, which also means a lot of biases are introduced.
As we can see, using only temporal loss yields us a counter-intuitive result, a $1.00$ probability for both the object on the left of the camera and on the right of the camera.
This is mainly due to the imbalanced low-level supervision that is introduced by human bias.
By adding a contrastive loss, we can see an improvement in that the model predicts the ground truth position with a higher probability compared to its counterpart.
Incorporating a violation loss can further improve the performance in that the sum of the probabilities is closer to $1.0$.

\begin{figure*}[tb]
    \centering
    \includegraphics[width=\linewidth]{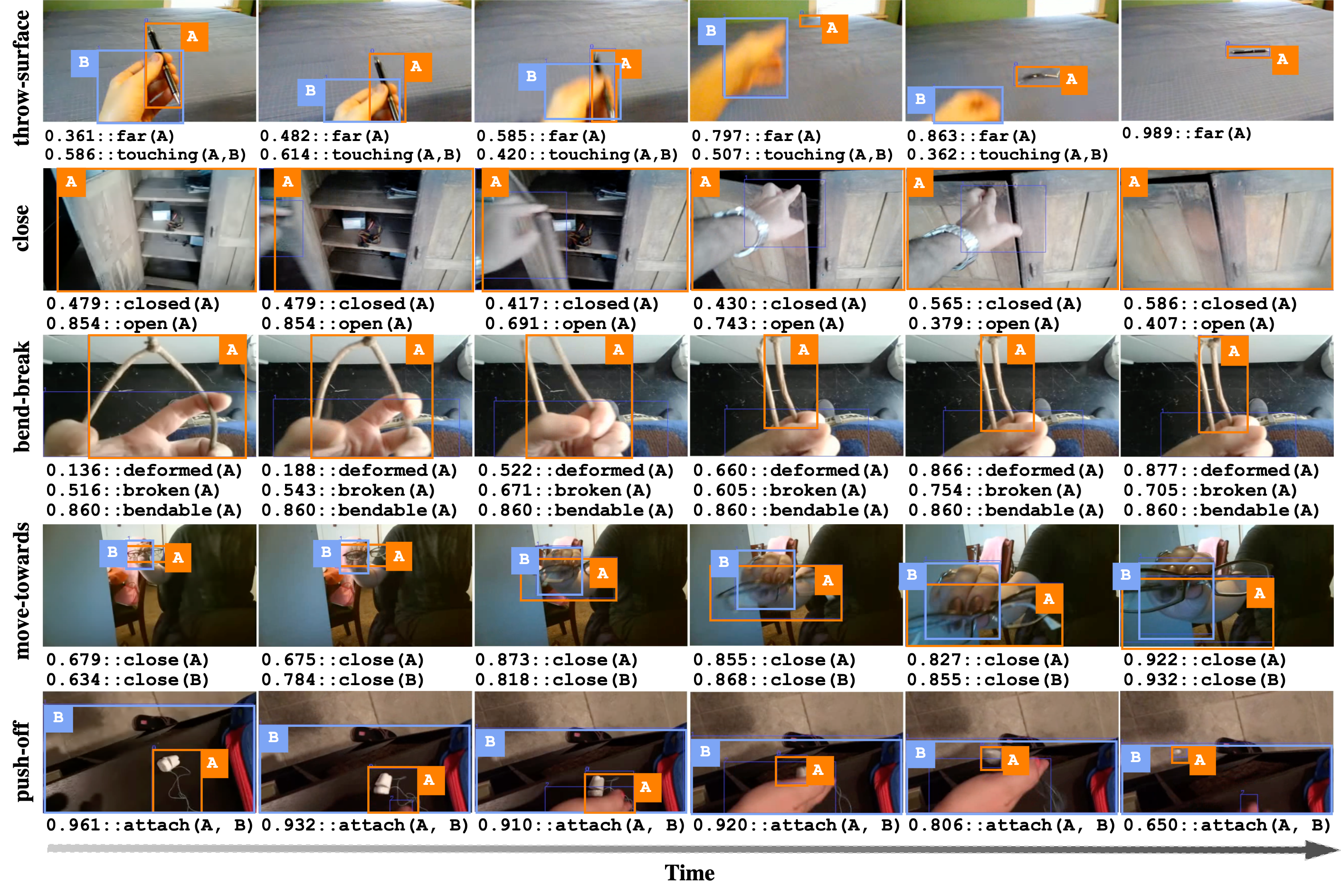}
    \caption{
    The STSGs predicted by \ours\ on the 20BN dataset.
    The actions are labeled on the left.
    }
    \label{fig:20bn_qualitative_full}
\end{figure*}

\textbf{More Qualitative Studies}
We include more qualitative examples predicted by the \ours-P model in \figref{fig:20bn_qualitative_full}.

% \section{Temporal Reasoner}
% \label{app:temp_reasoner}
% We showcase one temporal reasoner design in this section that enables all linear temporal language features: global ($\square$), finally ($\lozenge$), next($\bigcirc$), and until($\mathbf{U}$).
% % Note that we only lay out the core matching process of LTL, while the reasoner can be configured and extended with all other Datalog features, such as grounding free variables. 
% This is a generic implementation that suits for all three datasets, and extendable to other temporal required scenarios. 

% \input{figures/datalog_implementation}

% \pagebreak
% \input{appendix/3_checklist}

%%%%%%%%%%%%%%%%%%%%%%%%%%%%%%%%%%%%%%%%%%%%%%%%%%%%%%%%%%%%

\end{document}